\documentclass{article} % For LaTeX2e
\usepackage{main,times}

\usepackage{amsmath,amsfonts,bm}

\def\eqref#1{equation~\ref{#1}}
\def\1{\bm{1}}

\DeclareMathAlphabet{\mathsfit}{\encodingdefault}{\sfdefault}{m}{sl}
\SetMathAlphabet{\mathsfit}{bold}{\encodingdefault}{\sfdefault}{bx}{n}

\usepackage{hyperref}
\usepackage{url}

\usepackage{multirow}
\usepackage{booktabs}
\usepackage[table]{xcolor} % 放在导言区
\usepackage{adjustbox}        % 表格整体缩放
\usepackage{array}            % 支持 p{宽度} 自动换行
\usepackage{tabularx}
\usepackage{amsfonts}
\usepackage{subcaption}
\usepackage{wrapfig}
\newcommand{\method}{Glance Forcing}

\title{\method: Making Bidirectional Diffusion Autoregressive at a Glance}

\iclrfinalcopy

\author{%
  Zhuobai Dong$^{1}$, Junchao Yi$^{2}$, Hu Jian Guo$^{3}$, 
  \textbf{Linjie Li}$^{4}$, 
  \textbf{Rui Zhao}$^{5}$\thanks{Corresponding author.},
  \textbf{Alex Jinpeng Wang}$^{6}$ \\
  $^{1}$Wuhan University \quad $^{2}$University of Electronic Science and Technology of China \\
  $^{3}$Sun Yat-sen University \quad $^{4}$University of Washington \\
  $^{5}$National University of Singapore \quad $^{6}$Central South University
}

\begin{document}

% \maketitle

\pagestyle{empty} % 隐藏当前页页眉页脚（或使用 \thispagestyle{plain}）
\maketitle
\thispagestyle{empty} % 确保首页不显示顶部页眉

\begin{figure}[ht]

\vspace{-0.8cm}
\centering
\includegraphics[width=\columnwidth]{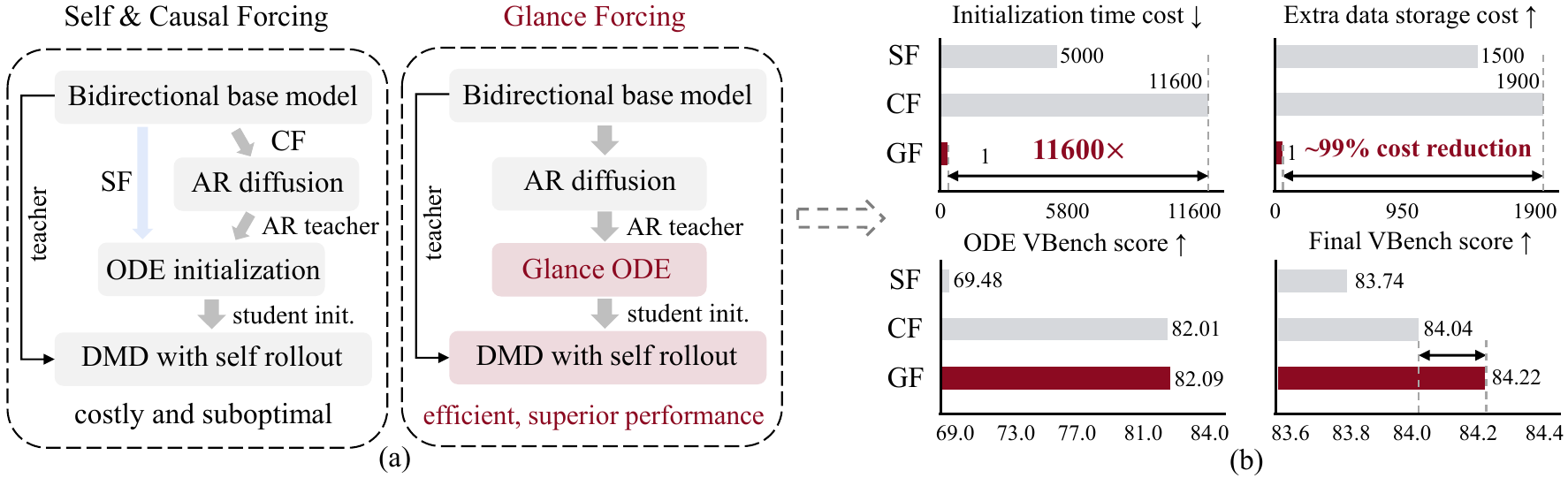}
\vspace{-0.7cm}
\caption{\textbf{Overall framework of our \method{} and the comparison with existing methods.} 
(a) Causal Forcing (CF) fixes Self Forcing (SF)’s architecture gap but remains costly; our \method{} (GF) is efficient, and superior. (b) Our GF reduces training cost by 11600×, requires only 1 extra sample curation, and achieves higher VBench scores than CF.}
\label{teaser}
\vspace{-0.1cm}
\end{figure}

\begin{abstract}
Autoregressive generation underlies real-time, causal video synthesis and is the backbone of video world models. Today's dominant approach distills a bidirectional diffusion model into a causal few-step student, but existing pipelines make this conversion prohibitively expensive, requiring offline \textit{Probability Flow ODE} (PF-ODE) trajectories that cost tens of thousands of GPU-hours and terabytes of storage. We show it can be made dramatically more efficient.
Pushing this efficiency to its extreme, we reduce trajectory supervision to a single sample, the strictest setting possible, and find that it causes severe quality collapse.
We trace this to the teacher's trajectory geometry: the high-noise regime, governing semantic layout and motion, and the low-noise regime, governing fine texture exhibit fundamentally different curvature and optimization behavior, so a single monolithic student cannot jointly fit both, causing severe representation interference.
Motivated by this, we propose \method, a lightweight distillation framework that decouples the denoising trajectory into two complementary slow and fast experts, each guided by a tailored, stage-specific objective, while keeping the pretrained teacher entirely frozen.
This decoupling further reveals a general mechanism underlying few-step distillation: PF-ODE distillation mainly straightens the few-step trajectory geometry, while the subsequent distribution-matching stage independently governs distributional quality.
Using only one PF-ODE trajectory, \method{} already surpasses the fully-trained state-of-the-art baseline, raising the VBench Total score from 84.04 to 84.22 while cutting trajectory-collection cost from thousands of GPU-hours to about \textbf{one A100-GPU-hour}, with markedly larger gains in dynamic degree, VisionReward, and instruction-following, and consistent preference from human raters over strong baselines.
HomePage: \url{https://zhuobaidong.github.io/Glance-Forcing/}
\end{abstract}

\vspace{-0.4cm}
\section{Introduction}
\vspace{-0.1cm}
% background
% video diffusion model and ar diffusion model

Diffusion-based video generation has achieved remarkable progress, with recent models such as Sora~\citep{openai2024sora}, Wan~\citep{wan2025wan}, and Seedance~\citep{seedance2026seedance} producing videos with high visual fidelity and complex spatiotemporal dynamics. 
However, their bidirectional denoising paradigm processes the entire video jointly, making inference increasingly expensive as video length grows.
Autoregressive (AR) diffusion models~\citep{jin2024pyramidal,chen2024diffusion,yin2025slow} address this limitation by performing causal rollout across frames or chunks while retaining diffusion-based generation within each
segment.
Recent AR diffusion distillation methods~\citep{huang2026self,zhu2026causal} have achieved promising results by
distilling bidirectional diffusion models into few-step AR students.
However, as shown in Fig.~\ref{teaser}(b),
Despite these advances, existing AR distillation pipelines typically rely on expensive offline \textit{Probability Flow ODE} (PF-ODE) trajectories as supervision. Constructing and storing such trajectories over large video datasets introduces substantial overhead, creating a severe preprocessing bottleneck.

\begin{figure*}[t]
\centering
\includegraphics[width=1\textwidth]{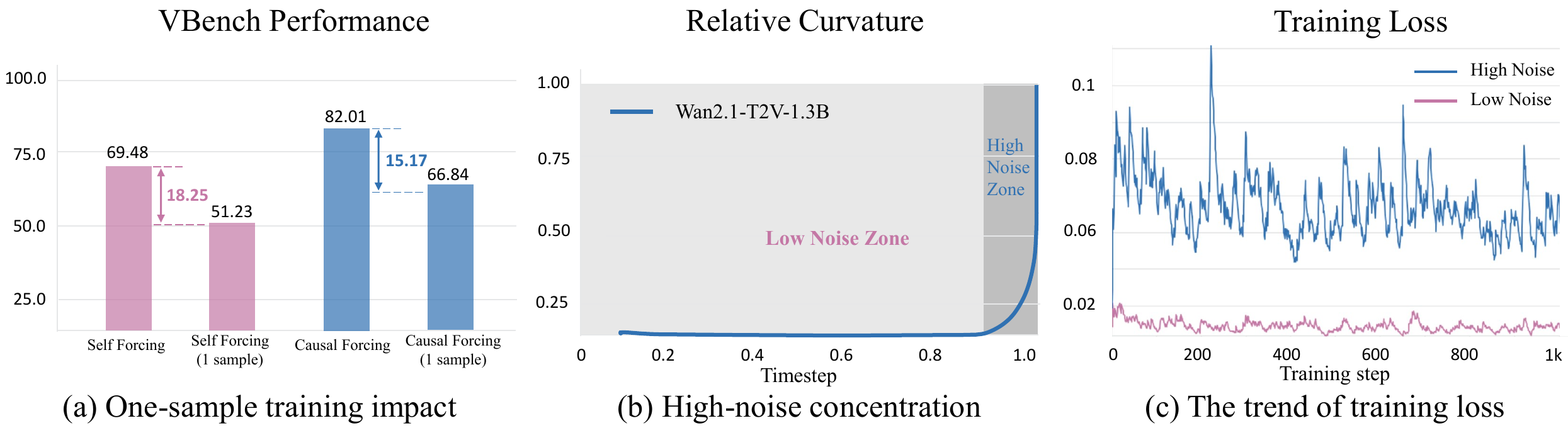}
\vspace{-0.6cm}
\caption{\textbf{Motivation and Trajectory Analysis.} (a) VBench performance drops significantly for both forcing methods when trained on a single sample on the ODE stage. (b) Relative trajectory-curvature profiles show high-noise concentration for Wan video generation. (c) Distillation loss exhibit significant differences between samples with high and low noise levels.}
\label{fig11}
\vspace{-0.6cm}
\end{figure*}

% Motivation

A natural way is therefore to reduce the amount of trajectory supervision. However, as shown in Fig.~\ref{fig11}(a), aggressively reducing the training data to one sample causes severe performance degradation. This raises a natral question: \textit{\textbf{why does few-step AR diffusion struggle under extremely sparse trajectory supervision?}} We find that the diffusion process exhibit an intriguing temporal dynamic during generation. As illustrated in Fig.~\ref{fig11}(b), more than 92.5\% of the cumulative trajectory curvature is concentrated in the early, high-noise regime. Furthermore, the training losses in Fig.~\ref{fig11}(c) exhibit markedly different optimization behaviors between the two regimes.
Under extremely sparse supervision, particularly in the one-sample regime, forcing a single monolithic network to capture these two conflicting learning regimes leads to severe representation interference.

% method
\textbf{\textit{The key insight is simple: different denoising phases require different experts.}}
Based on this observation, we present \method{}, a phase-aware distillation framework designed for extremely sparse-data training. Instead of constructing costly PF-ODE trajectories, \method{} learns directly from a single sample trajectory and decomposes the denoising process into two complementary experts using a dual-LoRA~\citep{hu2021lora} architecture: \textbf{Slow-LoRA} specializes in high-noise semantic trajectory synthesis, while \textbf{Fast-LoRA} focuses on low-noise detail refinement. Since only LoRA parameters are optimized, our framework preserves the pretrained diffusion teacher model. 
We further introduce lightweight phase-specific objectives to capture their distinct characteristics, including a temporal frequency loss for stable motion synthesis and a spatial high-pass loss for local detail refinement, without introducing an additional student network.

Experiments demonstrate that \method{} consistently improves the quality and efficiency across scales. In the one-sample regime, \method{} outperforms Causal Forcing, improving the VBench total score from 84.04 to 84.22 while cutting trajectory-collection cost from thousands of GPU-hours to about one A100-hour. 
As the training budget scales to match baseline, \method{} delivers further gains, reaching a peak VBench total score of 84.33. 

The contributions are summarized as follows: (1) We find that PF-ODE distillation mainly straightens the few-step trajectory rather than modeling the data distribution, where a single trajectory already suffices, as long as it is properly exploited across different noise regimes. (2) We propose \method, a lightweight dual-LoRA framework that decouples denoising into a Slow expert for layout and motion, and a Fast expert for detail, each with a tailored objective. (3) Extensive experiments show that \method{} matches or surpasses other methods while cutting the training cost by orders of magnitude, with consistent gains as the training budget scales up.

% \input{sec/1_introduction}

% \vspace{-0.3cm}

\section{Related Works}

% \vspace{-0.2cm}
\noindent\textbf{Bidirectional and Autoregressive Video Generation.}\quad
Current video diffusion models primarily fall into two paradigms. \emph{Bidirectional} models denoise an entire clip using full spatiotemporal attention, achieving strong global coherence at the cost of quadratic complexity in sequence length~\citep{blattmann2023stable,blattmann2023align,ho2022videodiffusion,ho2022imagenvideo}. While highly effective for offline synthesis, such overhead makes them impractical for interactive scenarios that demand low per-frame latency.

Autoregressive (causal) video generators factorize the joint distribution into $p_\theta(x^{1:K}\mid c)=\prod_k p_\theta(x^k\mid x^{<k},c)$, allowing each generated block to serve as context for subsequent blocks via a Key-Value (KV) cache. This factorization naturally aligns with real-time streaming, as only the current block undergoes denoising while historical context remains static in the cache. 
To train such models, two canonical strategies are typically employed: Teacher Forcing (TF)~\citep{jin2024pyramidal} and Diffusion Forcing (DF)~\citep{chen2024diffusion}. TF models $p_{\text{data}}(\mathbf{x}_0^i \mid \mathbf{x}_0^{<i})$ conditioned on a clean prefix of past frames.
Conversely, DF targets the noise-conditioned distribution $p_{\text{DF}}(\mathbf{x}_0^i \mid \mathbf{x}_t^{<i})$, where noise is independently injected into prior frames via $q_{t|0}(\mathbf{x}_t^{<i}\mid \mathbf{x}_0^{<i})$.

Recent advances have focused on closing the gaps in causal diffusion training. Self Forcing~\citep{huang2026self} first demonstrated that training on self-generated context with a holistic video-level loss effectively mitigates the train--test discrepancy. Causal Forcing~\citep{zhu2026causal} further showed that initializing ODE paths with an autoregressive teacher provably bridges the architectural gap between full and causal attention. Subsequent systems, including LongLive~\citep{yang2025longlive}, Rolling Forcing~\citep{liu2026rollingforcing}, and Self-Forcing++~\citep{cui2025selfforcingpp}, have pushed this boundaries further.
However, these methods heavily rely on constructing dense PF-ODE trajectories collected over large datasets, incurring prohibitive compute and storage costs.

\noindent\textbf{Diffusion Distillation.}\quad
Diffusion distillation techniques generally fall into two categories for compressing multi-step diffusion or flow models into few-step generators. 
The first family leverages continuous-time transport trajectories connecting noise and data distributions. 
Flow Matching~\citep{lipman2022flow} and Rectified Flow Transformers~\citep{esser2024scaling} learn velocity fields parameterizing these transport trajectories, whereas Consistency Distillation (CD)~\citep{song2023consistency,song2023improved} parameterizes a flow map $G_\theta: (\mathbf{x}_t,t) \mapsto \mathbf{x}_0$ that directly maps any noisy state $\mathbf{x}_t$ to the clean endpoint $\mathbf{x}_0$ along the teacher's PF-ODE. 
Recent interactive video generation systems~\citep{yin2025slow,huang2026self,zhu2026causal} often adopt a simplified variant of CD, optimizing the student model $G_\theta$ via direct regression:
\begin{equation}
\theta^* = \arg\min_\theta \mathbb{E}_{t,\mathbf{x}_t} \left[ \| G_\theta(\mathbf{x}_t, t) - \mathbf{x}_0 \|^2 \right],
\end{equation}
where $\mathbf{x}_t$ and $\mathbf{x}_0$ are sampled from the teacher PF-ODE. However, this line of work implicitly assumes that the trajectories are sufficiently smooth to compress without severe distortion.

The second family, score distillation~\citep{wang2023prolificdreamer, luo2023diff}, distills multi-step teacher models into few-step students by directly matching the student's generative distribution $p_\theta (\tilde{\mathbf{x}})$ to the target data distribution. 
A prominent instantiation is Distribution Matching Distillation (DMD)~\citep{yin2024dmd,yin2024dmd2}. 
Instead of tracking specific trajectories, DMD estimates a reverse-KL gradient using the difference between a real-data score $s_{\mathrm{real}}$ and a learned fake-data score $s_\phi$, steering the student toward the target distribution.
Recent video extensions adapt DMD to autoregressive setups via windowed self-rolled sequences~\citep{nie2026tmd}, reward-weighted distribution matching~\citep{lu2025rewardforcing}, and diagonal multi-step scheduling~\citep{liu2026diagonaldistill}.
However, these methods neglect the intrinsic spatiotemporal dynamics of the video generation process, leading to suboptimal training.

\section{Method}
\subsection{Preliminary}
\noindent\textbf{Diffusion Model} learns data distributions by gradually transforming noise into data through a parameterized denoising process.
In the forward process, Gaussian noise is progressively added to clean data $\mathbf{x}_0 \sim p_{\mathrm{data}}(\mathbf{x}_0)$:
\begin{align}
q(\boldsymbol{x}_t|\boldsymbol{x}_0) = \mathcal{N}(\boldsymbol{x}_t;\sqrt{\alpha_t}\boldsymbol{x}_0,\sqrt{1-\alpha_t}\boldsymbol{I}),
\label{eq:forward}
\end{align}
where $\{\alpha_t\}_{t=1}^T$ controls the noise schedule. The reverse process, typically parameterized by a UNet or transformer $\boldsymbol{\epsilon}_{\theta}$, is trained to predict the noise:
\begin{align}
\mathcal{L}_{DM} = \mathbb{E}_{\boldsymbol{x}, \boldsymbol{\epsilon} \sim \mathcal{N}(0,1),t} \big[||\boldsymbol{\epsilon} - \boldsymbol{\epsilon}_{\theta}(\boldsymbol{x}_t,t)||^2_2\big].
\end{align}
During inference, a clean sample $\boldsymbol{x}_0$ can be recovered through iterative denoising:
\begin{align}
p(\boldsymbol{x}_{t-1}|\boldsymbol{x}_t) = \mathcal{N}(\boldsymbol{x}_{t-1};\mu_{\theta}(\boldsymbol{x}_t,t),\Sigma_{\theta}(\boldsymbol{x}_t,t)),
\end{align}
where $\mu_{\theta}$ and $\Sigma_{\theta}$ are learned parameters.

\noindent\textbf{Consistency Distillation} utilizes a pre-trained model $\boldsymbol{\epsilon}_{\theta}$ as the teacher $F_T$ to distill its knowledge into a student model $F_S$ initialized with $\boldsymbol{\epsilon}_{\theta}$, allowing for faster sampling with fewer steps. 
Specifically, consistency distillation trains the student model $F_S$ to directly map any point $\boldsymbol{x}_{t_n}$ on the trajectory of the ODE solver $\Phi$ to its endpoint $\boldsymbol{x}_{t_{end}}$. The learning objective is formulated as:
\begin{equation}
\begin{aligned}
\mathcal{L}_{CD} = \mathbb{E}_{\boldsymbol{x},t_n}||\Phi(\boldsymbol{x}_{t_{n}},F_S(\boldsymbol{x}_{t_{n}},t_n,c),t_{end})- \\ \Phi(\hat{\boldsymbol{x}}_{t_{n-1}},F_S^{-}(\hat{\boldsymbol{x}}_{t_{n-1}},t_{n-1},c),t_{end})||_2^2.
\end{aligned}
\end{equation}
where $F_S^-$ denotes the exponential moving average (EMA) of $F_S$, and $\hat{\mathbf{x}}_{t_{n-1}}$ is the adjacent state along the trajectory predicted by the teacher $F_T$:
\begin{equation}
\begin{aligned}
\hat{\boldsymbol{x}}_{t_{n-1}} = \Phi(\boldsymbol{x}_{t_{n}},F_T(\boldsymbol{x}_{t_{n}},t_n,c),t_{n-1}).
\end{aligned}
\end{equation}
Due to its stable training, Consistency distillation has garnered widespread attention and research.

\subsection{Why Can One Sample Suffice?}
\label{sec:one-sample}

The Causal Forcing~\citep{zhu2026causal} pipeline involves three stages: (1) multi-step AR diffusion training via teacher forcing; (2) causal ODE initialization with
the AR teacher; and (3) asymmetric DMD with student self-rollout.
While existing approaches typically use large-scale data for ODE initialization, our method uses only a single sample trajectory.

\begin{figure*}[t]
\centering
\includegraphics[width=1\textwidth]{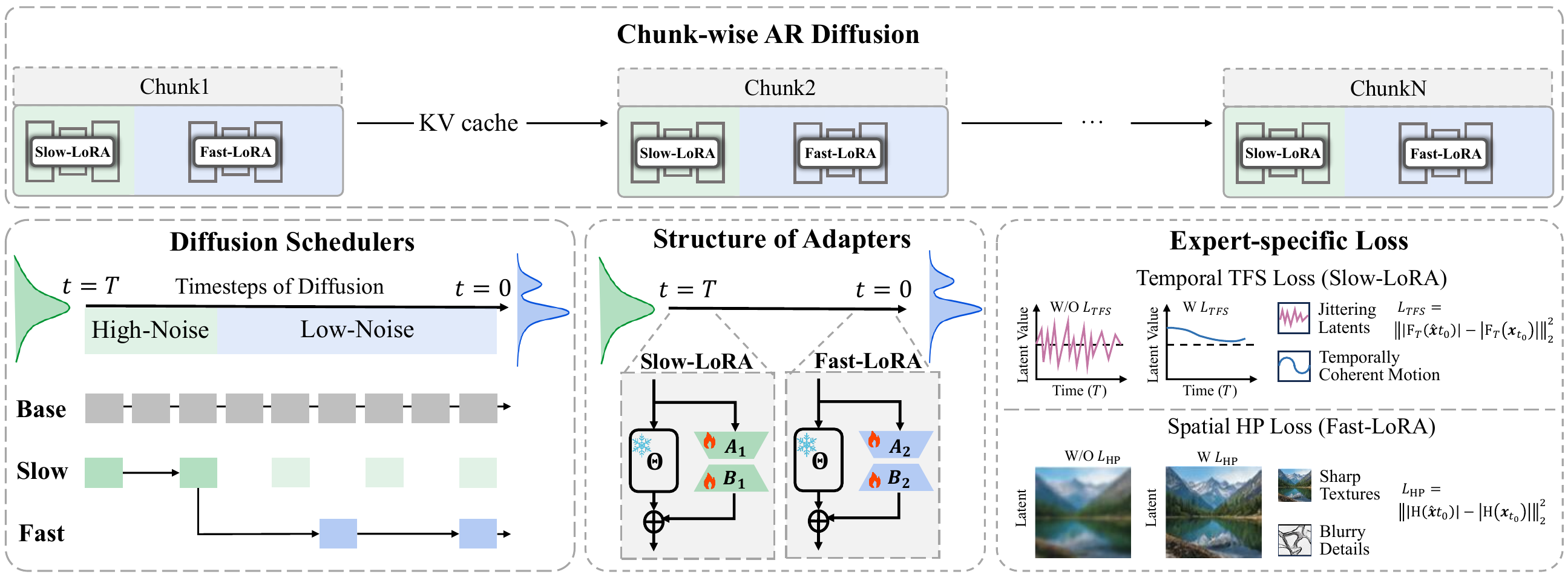}
\vspace{-0.7cm}
\caption{\textbf{Overview of the Slow-Fast paradigm.} We introduce a chunk-wise autoregressive diffusion framework, where each chunk is progressively denoised with stage-specific LoRA adapters. Slow-LoRA focuses on coarse structure generation in early stages, while Fast-LoRA refines details in later stages. Expert-specific objectives further enhance temporal consistency and spatial fidelity.}
\label{method}
\vspace{-0.5cm}
\end{figure*}

\noindent\textbf{Hypothesis.}\quad
We hypothesize that ODE initialization primarily serves to reshape the long-step diffusion trajectory into a coherent few-step flow, rather than fully modeling the target data distribution. Given the strong spatiotemporal priors of the AR teacher, its induced trajectories share substantial structural similarities across samples. Therefore, a small number of trajectories may already provide sufficient supervision to establish the few-step flow across the time domain, while the subsequent DMD stage can perform the necessary distributional refinement.

To examine this hypothesis, we vary the number of trajectories used for ODE initialization and evaluate the resulting denoising trajectories under 4-step inference. We compare models initialized with different amounts of ODE supervision and examine the effect of DMD on the learned trajectory.

\noindent\textbf{Trajectory Metrics.}\quad
We evaluate trajectory geometry via two metrics. \emph{Mean Trajectory Straightness} measures the ratio between the direct displacement and the accumulated path length:
\begin{equation}
\mathcal{S} =
\frac{\lVert z_T-z_0\rVert_2}
{\sum_{t=1}^{T}\lVert z_t-z_{t-1}\rVert_2}.
\end{equation}
A value closer to $1$ indicates a straighter trajectory with less deviation from the overall flow direction. We further measure \emph{Step-wise Cosine Similarity} between each denoising update $v_t=z_{t-1}-z_t$ and the overall trajectory direction $V_{\mathrm{total}}=z_0-z_T$:
% \vspace{-3mm}
\begin{equation} \mathrm{CosSim}_t = \frac{v_t\cdot V_{\mathrm{total}}} {\lVert v_t\rVert_2\lVert V_{\mathrm{total}}\rVert_2}. \end{equation}
% \vspace{-5pt}
The two metrics capture global trajectory straightness and local directional consistency, respectively.

\begin{wrapfigure}{r}{0.40\textwidth} % r 表示置右，0.48\textwidth 大约是整个页宽的一半
  \vspace{-0.6cm}
  \centering
  \includegraphics[width=\linewidth]{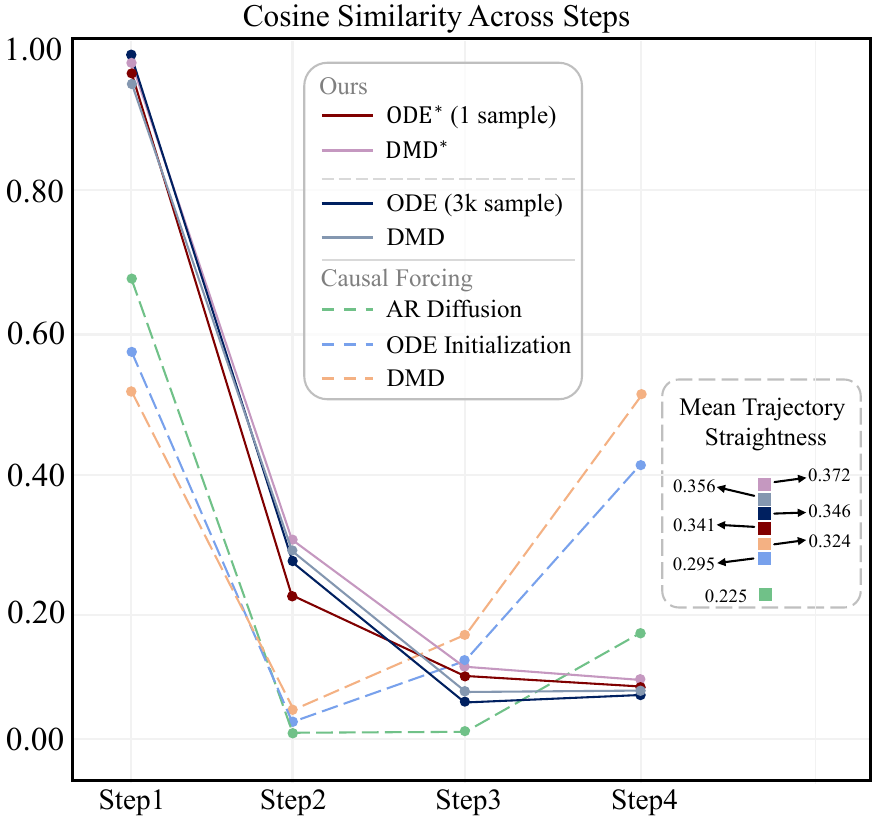}
  \vspace{-0.7cm}
  \caption{Comparison of cosine similarity trends and trajectory straightness across different models.}
  \label{trajectory}
  \vspace{-0.6cm}
\end{wrapfigure}

\noindent\textbf{Findings.}\quad
As shown in Fig.~\ref{trajectory}, ODE initialization substantially improves both trajectory straightness and step-wise directional alignment under 4-step inference compared with the long-step trajectory produced by the pretrained AR teacher. Crucially, scaling ODE supervision from a single trajectory to 3K yields only marginal changes in trajectory geometry. This shows that ODE-driven trajectory straightening depends primarily on covering the diffusion time domain rather than raw data volume. Moreover, subsequent DMD improves generation quality while leaving the overall trajectory geometry largely unchanged, indicating that it enhances output fidelity without altering the established flow dynamics.
These findings confirms that a single trajectory provides sufficient supervision for ODE to build a straight flow, while distributional refinement can be delegated to DMD.

\subsection{Suboptimal Trajectory Fitting in Consistency Distillation}
\label{suboptimal}

Sec.~\ref{sec:one-sample} shows that a single trajectory provides adequate geometric supervision for ODE initialization, but this does not guarantee that a single student can fit it well. We therefore turn from \emph{how much} supervision is needed to \emph{how hard} it is to fit: the central challenge is that a single student must approximate trajectory segments with substantially different geometric complexity across the diffusion time domain using one set of parameters. To quantify this complexity, we first measure the local deviation of the teacher trajectory from its global linear direction.

Inspired by Transition Matching Distillation~\citep{nie2026tmd}, we analyze trajectory non-linearity by measuring how much the teacher trajectory diverges from the straight chord connecting data and noise endpoints. Specifically, for adjacent states we quantify the local velocity deviation $C(t_i) = \frac{1}{d} \left\| \frac{x_{t_i}-x_{t_{i-1}}}{t_i-t_{i-1}}-(x_1-x_0) \right\|_2^2$, where $t=0$ and $t=1$ denote the clean data and high-noise endpoints, respectively, and $d$ is the latent dimension. Thus, $C(t_i)$ captures the per-coordinate squared error between local teacher velocity and the ideal linear trajectory.
As visualized in Fig.~\ref{fig11}(b), over $92.5\%$ of the total curvature mass is concentrated in the high-noise region ($t \ge 0.9$). 
This geometric discrepancy explains why standard ODE distillation with uniform timestep sampling is suboptimal: directly regressing highly curved, noisy intermediate states to clean endpoints creates an inherently inconsistent target.
Sampling details are provided in Appendix~\ref{training_details}.

% Furthermore, prior studies~\citep{anagnostidis2025flexidit} and empirical tracking of the synthesis process reveals distinct functional phases: early sampling stages undergo rapid state transitions to construct global semantic layouts and motion, whereas later stages refine local textures through subtle adjustments. 
% Consequently, high- and low-noise distillation pairs exhibit fundamentally different learning dynamics. 
% As illustrated by the consistency loss and gradient norm trends in Fig.~\ref{fig11}(b), forcing a single student model to simultaneously accommodate these conflicting tasks induces severe optimization friction, bottlenecking its representation capacity.

\begin{figure*}[t]
\centering
\includegraphics[width=1\textwidth]{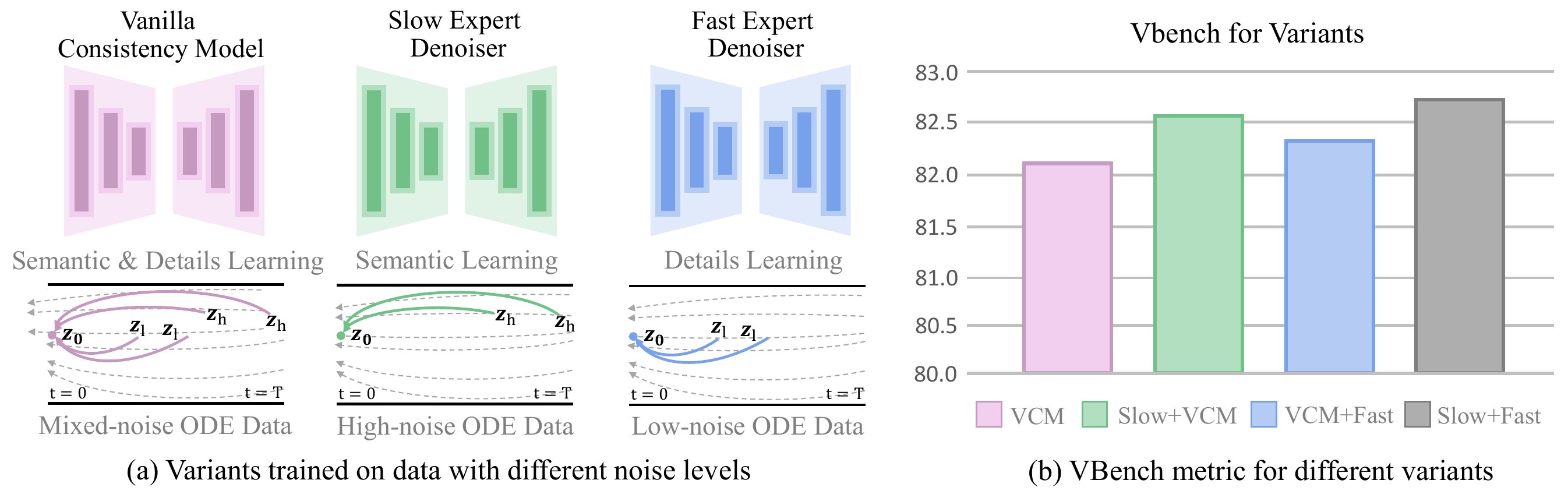}
\vspace{-0.7cm}
\caption{\textbf{Comparison of visual quality across denoiser variants trained on different noise levels}. Decoupling distillation into semantic and detail learning with two expert denoisers and combining them at inference yields the best quantitative visual results.}
\label{motivation}
\vspace{-0.5cm}
\end{figure*}

To validate our hypothesis, we conduct evaluations on the Causal Forcing framework. Specifically, we partition the ODE solution trajectory $(\mathbf{x}_{N}, \mathbf{x}_{N-1}, \dots, \mathbf{x}_{0})$ of the teacher model into two distinct sub-trajectories, demarcated by a threshold timestep $t_{\kappa}$ (we set $N=48$ and $\kappa=34$ by default).

The trajectory is divided based on distinct functional roles:
1) The high-noise regime ($\{\mathbf{x}_{t_{i}}\}_{i=\kappa}^{N}$) primarily governs the synthesis of global semantic layout and motion.
2) The low-noise regime ($\{\mathbf{x}_{t_{j}}\}_{j=0}^{\kappa}$) focuses on semantic refinement and fine-grained detail generation.

As illustrated in Fig.~\ref{motivation}(a), we train two specialized student models—a layout/motion expert $F_{\text{Slow}}$ and a detail expert $F_{\text{Fast}}$—to fit their respective sub-trajectories independently:
\begin{equation}
\begin{aligned}
\mathcal{L}_{\text{Slow}} = \mathbb{E}_{\boldsymbol{x},t_m\in[t_{\kappa},t_{N}]}||\Phi(\boldsymbol{x}_{t_{m}},F_{\text{Slow}}(\boldsymbol{x}_{t_{m}},t_m,c),t_{\kappa})- \\ \Phi(\hat{\boldsymbol{x}}_{t_{m-1}},F_{\text{Slow}}^{-}(\hat{\boldsymbol{x}}_{t_{m-1}},t_{m-1},c),t_{\kappa})||_2^2,
\end{aligned}
\label{eq:S1}
\end{equation}
\begin{equation}
\begin{aligned}
\mathcal{L}_{\text{Fast}} = \mathbb{E}_{\boldsymbol{x},t_n\in [t_{0},t_{\kappa}]}||\Phi(\boldsymbol{x}_{t_{n}},F_{\text{Fast}}(\boldsymbol{x}_{t_{n}},t_n,c),t_{0})- \\ \Phi(\hat{\boldsymbol{x}}_{t_{n-1}},F_{\text{Fast}}^{-}(\hat{\boldsymbol{x}}_{t_{n-1}},t_{n-1},c),t_{0})||_2^2.
\end{aligned}
\label{eq:S2}
\end{equation}
During inference, the two experts are dynamically switched according to the sampling stage. To isolate the contribution of each expert, we compare four configurations: \textit{vanilla VCM, Slow + VCM, VCM + Fast, and Slow + Fast}. 
As shown in Fig.~\ref{motivation}(b), $F_{\text{Slow}}$ and $F_{\text{Fast}}$ consistently outperform the vanilla VCM baseline in modeling structural semantics and fine details, respectively. These empirical findings strongly support our hypothesis, confirming that decoupling trajectory optimization effectively alleviates the capacity bottleneck of training a unified model for conflicting objectives.

\subsection{Expert-specific Optimization Objective}

In addition to the consistency objectives, we introduce specialized expert loss functions for the semantic expert denoiser $F_{\text{Slow}}$ and the details expert denoiser $F_{\text{Fast}}$, as illustrated in Fig.~\ref{method}.

\noindent\textbf{Temporal Frequency Spectrum Loss.}
To improve the coherence of $F_{\text{Slow}}$, we introduce $\mathcal{L}_{TFS}$. Instead of operating solely in the time domain, $\mathcal{L}_{TFS}$ regularizes the temporal energy distribution in the frequency domain via a 1D discrete Fourier transform (1D-DFT) along the temporal axis $T$:
\begin{equation}
\begin{aligned}
\hat{\boldsymbol{x}}_{t_0} &= \Phi(\boldsymbol{x}_{t_{m}}, F_{\text{Slow}}(\boldsymbol{x}_{t_{m}}, t_m, c), t_{0}), \\
\mathcal{L}_{TFS} &= \left\| \left| \mathcal{F}_{1\text{D-DFT}}(\hat{\boldsymbol{x}}_{t_0}) \right| - \left| \mathcal{F}_{1\text{D-DFT}}(\boldsymbol{x}_{t_0}) \right| \right\|_2^2,
\end{aligned}
\end{equation}
where $\boldsymbol{x}_{t_0}$ denotes the target ODE latent, $\hat{\boldsymbol{x}}_{t_0}$ denotes the latent estimated by $F_{\text{Slow}}$, and $|\mathcal{F}_{1\text{D-DFT}}(\cdot)|$ denotes the magnitude of the complex frequency spectra. By aligning the temporal spectral energy, $\mathcal{L}_{TFS}$ effectively suppresses high-frequency temporal flickering while preserving motion dynamics.

\noindent\textbf{Spatial High-Pass Loss.}
To guide $F_{\text{Fast}}$ in synthesizing sharp textures and structural details, we introduce $\mathcal{L}_{HP}$, which penalizes structural discrepancies via a fixed 2D Laplacian operator:
\begin{equation}
\begin{aligned}
\hat{\boldsymbol{x}}_{t_0} &= \Phi(\boldsymbol{x}_{t_{n}}, F_{\text{Fast}}(\boldsymbol{x}_{t_{n}}, t_n, c), t_{0}), \\
\boldsymbol{H}(\boldsymbol{x}) &= \text{Conv2D}(\boldsymbol{x}, \boldsymbol{K}_{\text{Laplacian}}), \\
\mathcal{L}_{HP} &= \left\| \boldsymbol{H}(\hat{\boldsymbol{x}}_{t_0}) - \boldsymbol{H}(\boldsymbol{x}_{t_0}) \right\|_2^2,
\end{aligned}
\end{equation}
where $\boldsymbol{K}_{\text{Laplacian}}$ is a non-trainable, channel-wise 2D Laplacian kernel that extracts spatial high-frequency edges. By directly optimizing the high-pass filtered latent space, $\mathcal{L}_{HP}$ encourages crisp boundaries and fine textures without compromising the underlying semantic structure.

\section{Experiments}
\subsection{Setup}
\noindent\textbf{Implementation details.}\quad
\label{implementation}
Following prior work~\citep{huang2026self,zhu2026causal}, we adopt Wan2.1-1.3B~\citep{wan2025wan} as the base model for our AR few-step generator, producing $480 \times 832$ videos with 81 frames via chunk-wise autoregressive generation.
Our method builds upon the three-stage distillation pipeline of Causal Forcing.
All training is conducted on $8 \times$ A100 GPUs.
The default single training sample is shown in Fig.~\ref{one sample}.
% , while inference is performed on H100.
Additional details are provided in Appendix~\ref{training_details}.

\noindent\textbf{Evaluation.}\quad
Following Causal Forcing, we evaluate on VBench~\citep{huang2024vbench} and VisionReward~\citep{xu2024visionreward}.We report the overall scores of both benchmarks, and additionally evaluate dynamic degree and instruction following using the 100 prompts from Causal Forcing. All metrics are scaled by 100. We additionally report first-frame latency and throughput, measured on a single H100 GPU, excluding VAE-related time.

\subsection{Main Results}
\begin{table*}[t]
\footnotesize
\setlength{\tabcolsep}{3.5pt} % 缩小列间距（默认通常是 6pt）
\caption{\textbf{Comparisons with existing methods.} Our method achieves generation quality comparable to or better than previous SOTA methods. Data, Dyn., Vision., and Instruct. denote Training Data, Dynamic Degree, VisionReward, and Instruction Following, respectively. * indicates the same set of models across stages. Throughput and latency are measured in FPS and seconds on H100 GPU.}
\vspace{-0.4cm}
\label{table1}
\centering
\begin{tabular*}{\textwidth}{@{} p{2.35cm} @{\extracolsep{\fill}} ccccccccc @{}}
    \toprule
    Model & Data & Thru. $\uparrow$ & Latency $\downarrow$ & Total $\uparrow$ & Quality $\uparrow$ & Semantic $\uparrow$ & Dyn. $\uparrow$ & Vision. $\uparrow$ & Instruct. $\uparrow$ \\
    \midrule
    \multicolumn{10}{c}{\textit{Bidirectional Video Diffusion Models}} \\
    \midrule
    Wan2.1-1.3B         & - & 0.78 & 103 & 83.37 & 84.30 & 79.65 & 61 & 5.275 & 42 \\
    LTX-1.9B         & - & 8.98 & 13.5 & 79.83 & 81.88 & 71.62 & 46 & -6.218 & -38 \\
    \midrule
    \multicolumn{10}{c}{\textit{Autoregressive Video Diffusion Models}} \\
    \midrule
    SkyReels-V2   & - & 0.49 & 112 & 81.97 & 83.96 & 74.01 & 37 & 3.584 & 32 \\
    MAGI-1            & - & 0.19 & 282 & 78.88 & 81.67 & 67.72 & 42 & 0.773 & 8 \\
    \midrule
    \multicolumn{10}{c}{\textit{Initialized Autoregressive Video Models}} \\
    \midrule
    CausVid         & 1k & 17.0 & 0.69 & 66.58 & 70.42 & 51.23 & 72 & -13.724 & -60 \\
    Self Forcing   & 16k & 17.0 & 0.69 & 69.48 & 73.47 & 53.53 & \textbf{80} & -12.591 & -50 \\
    Causal Forcing & 3k & 17.0 & 0.69 & 82.01 & 82.59 & 79.73 & 64 & 0.603 & 50 \\
    \textbf{Glance Forcing}       
    & 3k & 16.4 & 0.75 & \textbf{82.69} & 83.33 & \textbf{80.11} & 63 & \textbf{2.310} & \textbf{52} \\
    \textbf{Glance Forcing*}       
    & 1 & 16.4 & 0.75 & 82.09 & \textbf{83.68} & 75.72 & 70 & 0.489 & 30 \\
    \midrule
    \multicolumn{10}{c}{\textit{Distilled Autoregressive Video Models}} \\
    \midrule
    CausVid          & 25k  & 17.0 & 0.69 & 81.33 & 83.98 & 70.72 & 62 & 5.741 & 12 \\
    Self Forcing    & 25k  & 17.0 & 0.69 & 83.74 & 84.48 & 80.77 & 57 & 5.820 & 48 \\
    Causal Forcing  & 25k   & 17.0 & 0.69 & 84.04 & 84.59 & \textbf{81.84} & 68 & 6.326 & 56 \\
    \textbf{Glance Forcing}       
    & 25k & 16.4 & 0.75 & \textbf{84.33} & 85.01 & 81.64 & \textbf{78} & \textbf{6.921} & \textbf{69} \\ 
    \textbf{Glance Forcing*}       
    & 25k & 16.4 & 0.75 & 84.22 & \textbf{85.50} & 79.11 & 75 & 6.870 & 64 \\
    \bottomrule
\end{tabular*}
\vspace{-0.6cm}
\end{table*}

We evaluate \method{} against representative video generation methods spanning three paradigms: bidirectional diffusion models (Wan2.1-1.3B~\citep{wan2025wan} and LTX-1.9B~\citep{hacohen2024ltx}), autoregressive diffusion models (SkyReels-V2-1.3B~\citep{chen2025skyreels} and MAGI-1-4.5B~\citep{teng2025magi}), and distilled autoregressive video models, including CausVid~\citep{yin2025slow}, Self Forcing~\citep{huang2026self}, and Causal Forcing~\citep{zhu2026causal}. 

\noindent\textbf{Quantitative Comparison.}\quad
Tab.~\ref{table1} reports the quantitative comparison. Remarkably, with only one training sample, \method{} already achieves competitive or better performance than Causal Forcing across most metrics, obtaining a Total score of 82.09 and consistently higher Dynamic, VisionReward, and Instruction Following scores. This demonstrates the strong data efficiency of our training strategy.
When scaling the training budget to match Causal Forcing, \method{} further improves the generation quality, achieving the best overall Total score of 82.69 among initialized autoregressive models. After autoregressive distillation, \method{} reaches 84.58, with further gains in Dynamic, VisionReward, and Instruction Following, showing that the benefits of phase-aware optimization persist and become more pronounced as the training budget scales.

\begin{figure*}[t]
\centering
\includegraphics[width=1\textwidth]{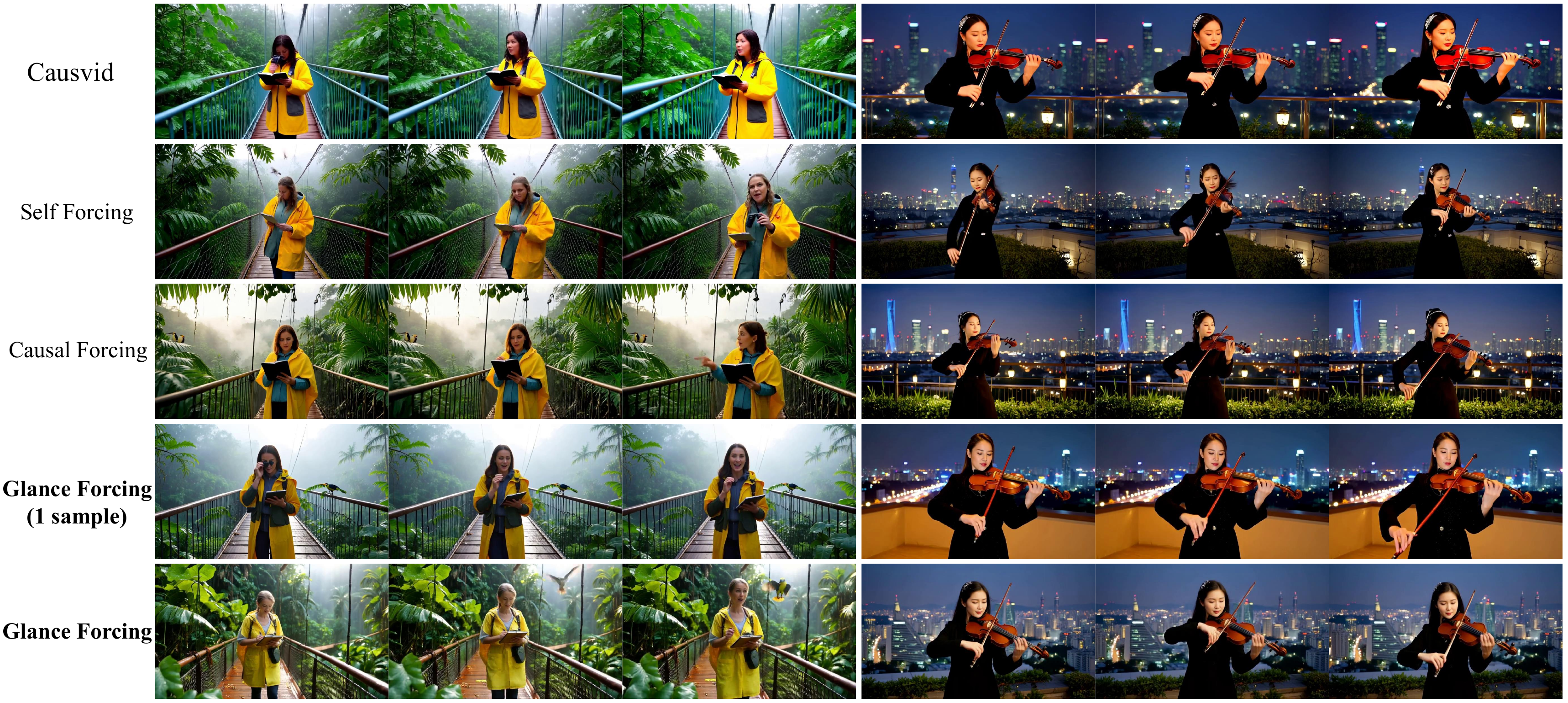}
\vspace{-0.6cm}
\caption{\textbf{Qualitative comparisons with existing methods.} Our method achieves competitive visual quality than state-of-the-art distilled autoregressive video models.}
\label{performance-comparison}
\vspace{-0.3cm}
\end{figure*}

\noindent\textbf{Qualitative Comparison.}\quad
Fig.~\ref{performance-comparison} provides qualitative comparisons that further support our quantitative results. 
With one training sample, \method{} produces visually competitive videos, with more stable object appearance, motion, and fine-grained details than existing distilled autoregressive models in several cases.
When scaling the training budget to match baseline Causal Forcing, \method{} further improves visual quality, producing more temporally consistent frames with sharper textures and clearer scene details.

\begin{table*}[t]
\vspace{-0.1cm}
\centering
% ===== 左侧：Table 2 =====
\begin{minipage}[t]{0.51\textwidth}
    \centering
    % \caption{\textbf{Human preference for \method{} against each baseline.} Counts aggregate votes from three annotators on 50 prompts (up to 150 votes per comparison). ``Win rate'' is the share of decided votes in which \method{} is preferred.}
    \caption{\textbf{Human preference for \method{} over each baseline.} Results aggregate votes from three annotators on 50 prompts, with 150 votes per comparison. ``Win rate'' denotes the fraction of decided votes in which \method{} is preferred.}
    \label{user}
    \small
    \setlength{\tabcolsep}{3pt}
    \begin{tabular}{l c c c c}
        \toprule
        \textbf{Baseline} & \textbf{Ours wins} & \textbf{wins} & \textbf{Total} & \textbf{Win rate} \\
        \midrule
        Self Forcing  & 137 & 13 & 150 & \textbf{91.3\%} \\
        Causal Forcing & 95 & 55 & 150 & \textbf{63.3\%} \\
        \bottomrule
    \end{tabular}
\end{minipage}% <--- 注意这里的百分号 %，防止产生额外空格
\hfill
% ===== 右侧：Table 3 =====
\begin{minipage}[t]{0.46\textwidth}
    \centering
    \caption{Ablation study of different components in \method{}.}
    \label{abatable}
    \vspace{-0.7em}
    \small
    \setlength{\tabcolsep}{3.5pt}
    \begin{tabularx}{\linewidth}{c c c c | c c c}
        \toprule
        \textbf{\#} & \multicolumn{3}{c|}{\textbf{Variants}} & \multicolumn{3}{c}{\textbf{VBench}} \\ 
        & SF & TFS & HP & Total & Quality & Semantic \\ 
        \midrule
        (1) &            &            &            & 83.73 & 84.32 & 81.37 \\
        (2) & \checkmark &            &            & 84.05 & 84.71 & 81.41 \\
        (3) & \checkmark & \checkmark &            & 84.20 & 84.76 & \textbf{81.96} \\
        (4) & \checkmark &            & \checkmark & 84.21 & 84.88 & 81.53 \\
        (5) & \checkmark & \checkmark & \checkmark & \textbf{84.33} & \textbf{85.01} & 81.64 \\   
        \bottomrule
    \end{tabularx}
\end{minipage}
\vspace{-0.55cm}
\end{table*}

\noindent\textbf{User Study.}\quad 
We conduct a pairwise human preference study comparing
\method{} with Self Forcing and Causal Forcing.
As shown in Tab.~\ref{user}, \method{} is consistently preferred by human annotators,
achieving win rates of 91.3\% and 63.3\% over Self Forcing and Causal Forcing,
respectively. These results align with the VBench evaluation in
Tab.~\ref{table1}, further supporting the improved perceptual quality of
\method{}.
Detailed study protocols are provided in
Appendix~\ref{app:user-study}.

% To further evaluate the effectiveness of our method, we conducted a pairwise human preference study
% comparing Glance Forcing against two causal baselines: Self Forcing and Causal Forcing. Specifically, We randomly sample 50 text prompts and generate corresponding video pairs for each baseline comparison. To eliminate potential bias, the spatial placement of the videos is randomized using a fixed seed, and all prompt samples are finalized prior to evaluation.

% Each video pair is independently evaluated by three human annotators. Annotators view both clips side-by-side in a continuous loop and select the preferred video (or a tie) based on overall quality, temporal consistency, motion realism, and prompt adherence. In total, this yields $50 \times 3 \times 2 = 300$ individual ratings. Annotators view both clips auto-playing on loop in adjacent panels and answer

% \begin{quote}
% \emph{``Which video is better overall, considering visual quality, motion realism, temporal consistency, and prompt alignment?''}
% \end{quote}
% selecting \emph{left}, \emph{right}, or \emph{tie}.

% Table~\ref{user} summarizes votes from the three annotators. Compared with the two baselines, \method{} is clearly preferred: 91.3\% over Self Forcing (137/150 decided votes, with three abstentions) and 74.6\% over Causal Forcing (112/150). These large margins are consistent with the VBench results in Table~\ref{table1}, where \method{} improves over Self Forcing and Causal Forcing by 0.84 and 0.54 points, respectively. This suggests that VBench largely agrees with human preference.

\subsection{Ablation studies}
We conduct ablation studies on VBench to evaluate the core components of \method.

\begin{figure*}[ht]
\vspace{-0.6cm}
\centering
\includegraphics[width=1\textwidth]{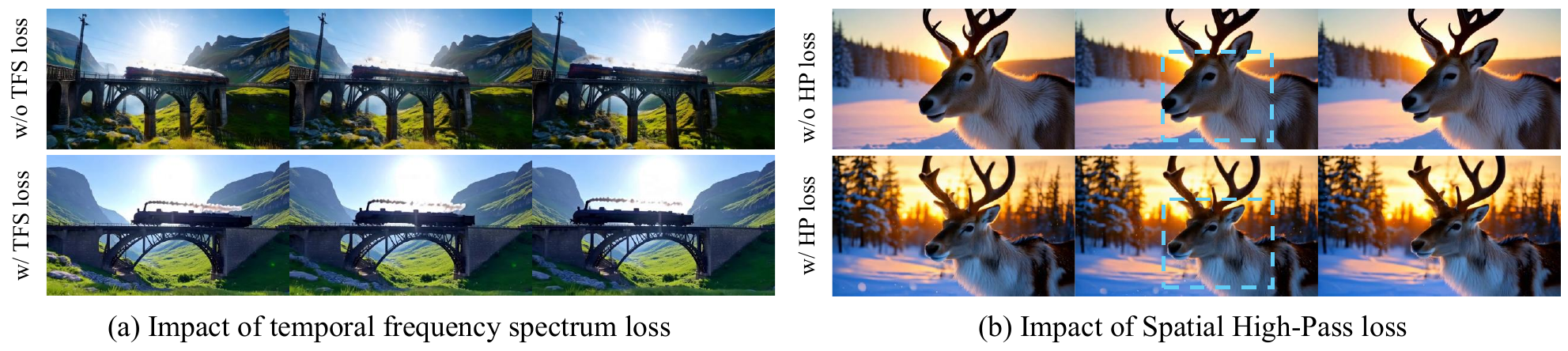}
\vspace{-0.7cm}
\caption{Ablation study Performance comparison.}
\label{abla}
\vspace{-0.5cm}
\end{figure*}

\noindent\textbf{Core Design.}\quad
As summarized in Tab.~\ref{abatable}, incorporating the Slow--Fast (SF) optimization decoupling improves the overall score from 83.98 to 84.33, demonstrating the necessity of separating coarse semantic synthesis from fine detail refinement. Integrating the Temporal Frequency Spectrum Loss (TFS) yields further quantitative gains, while the Spatial High-Pass Loss (HP) delivers additional improvements in both overall quality and spatial fidelity. 
Qualitatively, as illustrated in Fig.~\ref{abla}, TFS effectively enhances temporal consistency, whereas HP noticeably sharpens fine-grained spatial details. Together, these results confirm the synergistic efficacy of our core designs.

\noindent\textbf{Slow--Fast Switching.}\quad
We investigate different Slow--Fast step allocations within each 4-step chunk. 
As presented in Tab.~\ref{tab6}, our default S2--F2 consistently yields the optimal overall performance. Notably, deploying only Slow-LoRA (S4) yields higher visual quality, whereas Fast-LoRA (F4) better preserves motion dynamics and semantic consistency, revealing the complementary strengths of the two experts. 
In contrast, the imbalanced S1--F3 and S3--F1 schedules degrade performance. These results support an even allocation across the denoising trajectory as the best balance between semantic generation and detail refinement.

\subsection{Discussion}

\begin{figure}[t]
\centering
\includegraphics[width=\columnwidth]{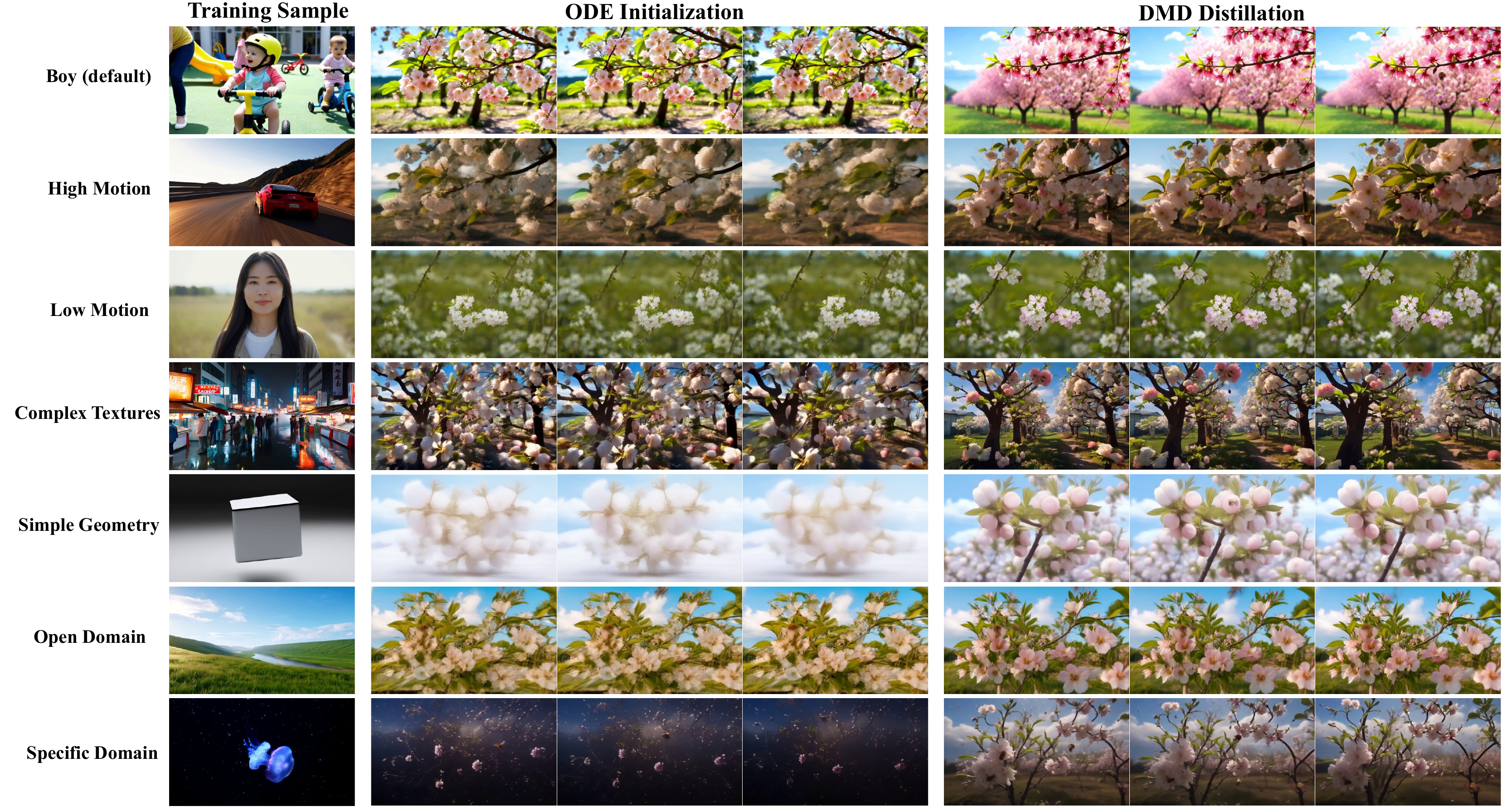}
\vspace{-0.7cm}
\caption{Qualitive results from one-sample training setting.}
\label{one sample}
\vspace{-0.4cm}
\end{figure}
\begin{table*}[t]
\centering
% ===== 左侧表格：Ablation study on 1 data sample =====
\begin{minipage}[c]{0.52\textwidth}
    \centering
    \caption{Ablation study on 1 sample.}
    \vspace{-0.4cm}
    \renewcommand{\arraystretch}{0.88} % 压缩行间距（默认是 1.0，可试 0.85~0.90）
    \label{tab:one sample}
    \resizebox{\linewidth}{!}{%
        \begin{tabular}{lcccccc}
        \toprule
        & \multicolumn{3}{c}{\textbf{ODE Initialization}}
        & \multicolumn{3}{c}{\textbf{DMD Distillation}} \\
        \cmidrule(lr){2-4} \cmidrule(lr){5-7}
        \textbf{Model}
        & \textbf{Total}
        & \textbf{Quality}
        & \textbf{Semantic}
        & \textbf{Total}
        & \textbf{Quality}
        & \textbf{Semantic} \\
        \midrule
        Boy (default)   & \textbf{82.09} & 83.68 & 75.72 & \textbf{84.22} & \textbf{85.50} & 79.11 \\
        \midrule
        \multicolumn{7}{c}{\textit{Motion Dynamics}} \\
        \midrule
        High Motion   & 75.37 & 74.91 & 77.23 & 82.72 & 83.64 & 79.04 \\
        Low Motion    & 78.23 & 78.14 & \textbf{78.59} & 79.81 & 80.02 & 78.98\\
        \midrule
        \multicolumn{7}{c}{\textit{Visual Complexity}} \\
        \midrule
        Complex Textures    & 78.10 & 78.02 & 78.42 & 82.65 & 83.56 & 79.02\\
        Simple Geometry & 73.79 & 74.36 & 71.51 & 83.52 & 84.51 & \textbf{79.56}\\
        \midrule
        \multicolumn{7}{c}{\textit{Domain Specificity}} \\
        \midrule
        Open Domain    & 79.45 & 80.03 & 77.14 & 83.48 & 84.47 & 79.52 \\
        Specific Domain    & 71.94 & 72.37 & 70.23 & 81.33 & 82.53 & 76.52\\
        \bottomrule
        \end{tabular}%
    }
\end{minipage}%
\hfill % 自动拉开左右距离
% ===== 右侧区域：上下叠加两个表格 =====
\begin{minipage}[c]{0.45\textwidth}
    \centering
    % 右上表格：Diffusion model for accelerating
    \caption{Diffusion model for accelerating.}
    \vspace{-0.4cm}
    \label{abla5}
    \resizebox{\linewidth}{!}{%
        \begin{tabular}{l c c c c c c}
            \toprule
            \textbf{Model} & \textbf{NFEs} & \textbf{Thru. $\uparrow$} & \textbf{Latency $\downarrow$} & \textbf{Total $\uparrow$} & \textbf{Quality $\uparrow$} & \textbf{Semantic $\uparrow$} \\
            \midrule
            Wan2.1        & 50 & 0.78 & 103  & 83.37 & 84.30 & 79.65 \\
            LTX-1.9B   & 20 & 8.98 & 13.5 & 79.83 & 81.88 & 71.62 \\
            \method                  & 8  & 17.0 & 1.38 & 82.16 & 83.42 & 77.12 \\
            \method                  & 4  & 34.0 & 0.69 & 81.14 & 82.27 & 76.65 \\
            \bottomrule
        \end{tabular}%
    }
    
    \vspace{0.8em} % 控制右上和右下两个表格之间的间距

    % 右下表格：Slowfast stage ablation study
    \caption{Slowfast stage ablation study.}
    \label{tab6}
    \vspace{-0.68em}
    \resizebox{\linewidth}{!}{%
        \begin{tabular}{lcccccc}
        \toprule
        \textbf{Model} & 
        \textbf{Total$\uparrow$} & 
        \textbf{Quality$\uparrow$} & \textbf{Semantic$\uparrow$} & \textbf{Dynamic$\uparrow$} & \textbf{Vision$\uparrow$}  & \textbf{Instruct$\uparrow$} \\
        \midrule
        S2–F2 & \textbf{84.33} & 85.01          & \textbf{81.64} & \textbf{78} & \textbf{6.921} & \textbf{69} \\
        S1–F3 & 83.92          & 84.65          & 81.00          & 76          & 2.865          & 41          \\
        S3–F1 & 84.01          & \textbf{85.05} & 79.85          & 68          & 4.210          & 55          \\
        S4       & 83.42          & 84.80          & 77.90          & 55          & 2.450          & 43          \\
        F4       & 83.15          & 84.20          & 78.95          & 72          & 1.310          & 31          \\
        \bottomrule
        \end{tabular}%
    }
\end{minipage}
\vspace{-0.4cm}
\end{table*}

\noindent\textbf{Systematic One-Sample Study.}\quad
We conduct three controlled one-sample studies across \textit{motion dynamics}, \textit{visual complexity}, and \textit{domain specificity}.
% We construct six representative samples along three dimensions: \textit{motion dynamics}, \textit{visual complexity}, and \textit{domain specificity}. Each dimension contains two contrasting samples under consistent video settings, camera trajectories, subject scales, and prompt structures. Specifically, High/Low Motion distinguish substantial from minimal temporal dynamics; Complex Textures/Simple Geometry contrast dense dynamic textures with clean geometric scenes; and Open/Specific Domain contrast broadly recognizable content with relatively rare or specialized content. This controlled categorization allows us to isolate different sources of generation difficulty while covering diverse visual scenarios. Detailed prompts are provided in Appendix~\ref{prompt}.
The results in Tab.~\ref{tab:one sample} reveal distinct sensitivity to sample characteristics. For ODE initialization, Low Motion and Open Domain achieve higher quality, whereas High Motion and Specific Domain exhibit larger degradation, suggesting that one-sample initialization is more sensitive to substantial temporal dynamics and specialized content. In contrast, DMD consistently recovers and improves generation quality across all samples, with particularly pronounced gains for High Motion and Simple Geometry. Notably, semantic scores remain relatively stable across settings, indicating that the primary variation induced by one-sample training lies in quality rather than semantic alignment.
Overall, a single sample provides effective supervision for both ODE and subsequent DMD refinement, although its effectiveness depends on characteristics of the selected sample. 
Detailed setup and prompts are provided in Appendix~\ref{prompt}.

\noindent\textbf{Generalization to Diffusion Acceleration.}\quad
We further validate \method{} beyond autoregressive generation by distilling the 50-step Wan2.1 model into an 8-step bidirectional diffusion model using a single training sample. As shown in Tab.~\ref{abla5}, the distilled model achieves comparable performance to Wan2.1 with over $5\times$ lower sampling cost, while maintaining modest degradation at 4 NFEs, demonstrating the generality of our phase-aware optimization.

\vspace{-0.3cm}
\section{Conclusion}

\vspace{-0.3cm}
We revisit the necessity of large-scale trajectory supervision for distilling autoregressive video diffusion models. We find that trajectory correction and distributional refinement play separable roles in the distillation pipeline, and that a single PF-ODE trajectory suffices once denoising is decoupled by noise regime. 
Built on this, \method{} uses a dual-LoRA design, with a Slow expert for layout/motion and a Fast expert for detail, each with a tailored objective.
Using one trajectory, \method{} matches or surpasses the baseline while reducing trajectory-collection cost from thousands of GPU-hours to about one A100-hour, with consistent gains as the training budget scales.

\clearpage

\section*{AI Use Statement}

In this work, we used generative AI tools to assist with translation during the drafting of the paper.
We have not used generative AI tools for generating synthetic data sets, developing theoretical models or conceptual frameworks, formulating mathematical claims, providing critical ingredients for proving mathematical claims, assisting in the writing of proofs, proposing or refining hypotheses, designing or providing feedback on research methodology or experiments, implementing methods, cleaning or reformatting datasets, supporting qualitative or thematic data analysis, and interpreting results, and these tasks are not applicable to this work.
Additionally, we used generative AI tools to edit the paper to improve readability.
We have reviewed all AI-assisted work: all AI-assisted text was checked and revised by the authors for accuracy and clarity.
We take responsibility for the final content of this work, including text, claims, or artifacts produced with the aid of generative AI.

\section*{Reproducibility}
Our models are built upon publicly available Wan model variants and datasets (OpenVid-1M and synthetic Wan videos). Training was conducted in a well-controlled environment, with all training details, hyperparameters, and implementation specifics provided in Section~\ref{implementation} and the Appendix~\ref{training_details} to ensure reproducibility.

% \section*{Ethics Statement}
% \label{sec:ethics}
% As video generative models become increasingly capable of producing realistic content that mimics world dynamics, we recognize the potential for misuse. While \method{} advances interactive content creation with intuitive controls, our approach naturally inherits potential risks from the underlying generative technology, including the creation of deceptive media. We emphasize the critical need for parallel development of safeguarding techniques such as watermarking, content authentication, and controlled access mechanisms. We encourage prioritizing responsible deployment strategies alongside capability improvements to ensure these tools benefit society while minimizing harms.

% \clearpage

\bibliography{main}

@String(CVPR= {IEEE Conf. Comput. Vis. Pattern Recog.})

@String(CVPR  = {CVPR})

@inproceedings{yin2025slow,
  title={From slow bidirectional to fast autoregressive video diffusion models},
  author={Yin, Tianwei and Zhang, Qiang and Zhang, Richard and Freeman, William T and Durand, Fredo and Shechtman, Eli and Huang, Xun},
  booktitle={Proceedings of the IEEE/CVF Conference on Computer Vision and Pattern Recognition},
  pages={22963--22974},
  year={2025}
}

@article{huang2026self,
  title={Self forcing: Bridging the train-test gap in autoregressive video diffusion},
  author={Huang, Xun and Li, Zhengqi and He, Guande and Zhou, Mingyuan and Shechtman, Eli},
  journal={Advances in Neural Information Processing Systems},
  volume={38},
  pages={167283--167308},
  year={2026}
}

@article{zhu2026causal,
  title={Causal Forcing: Autoregressive Diffusion Distillation Done Right for High-Quality Real-Time Interactive Video Generation},
  author={Zhu, Hongzhou and Zhao, Min and He, Guande and Su, Hang and Li, Chongxuan and Zhu, Jun},
  journal={arXiv preprint arXiv:2602.02214},
  year={2026}
}

@inproceedings{ho2022videodiffusion,
  title     = {Video Diffusion Models},
  author    = {Ho, Jonathan and Salimans, Tim and Gritsenko, Alexey and Chan, William and Norouzi, Mohammad and Fleet, David J.},
  booktitle = {Advances in Neural Information Processing Systems},
  volume    = {35},
  pages     = {8633--8646},
  year      = {2022},
  editor    = {Koyejo, S. and Mohamed, S. and Agarwal, A. and Belgrave, D. and Cho, K. and Oh, A.},
  publisher = {Curran Associates, Inc.},
  url       = {https://proceedings.neurips.cc/paper_files/paper/2022/file/39235c56aef13fb05a6adc95eb9d8d66-Paper-Conference.pdf}
}

@misc{ho2022imagenvideo,
  title         = {{Imagen Video}: High Definition Video Generation with Diffusion Models},
  author        = {Ho, Jonathan and Chan, William and Saharia, Chitwan and Whang, Jay and Gao, Ruiqi and Gritsenko, Alexey and Kingma, Diederik P. and Poole, Ben and Norouzi, Mohammad and Fleet, David J. and Salimans, Tim},
  year          = {2022},
  eprint        = {2210.02303},
  archivePrefix = {arXiv},
  primaryClass  = {cs.CV},
  doi           = {10.48550/arXiv.2210.02303},
  url           = {https://arxiv.org/abs/2210.02303}
}

@article{blattmann2023stable,
  title={Stable video diffusion: Scaling latent video diffusion models to large datasets},
  author={Blattmann, Andreas and Dockhorn, Tim and Kulal, Sumith and Mendelevitch, Daniel and Kilian, Maciej and Lorenz, Dominik and Levi, Yam and English, Zion and Voleti, Vikram and Letts, Adam and others},
  journal={arXiv preprint arXiv:2311.15127},
  year={2023}
}

@inproceedings{blattmann2023align,
  title={Align your latents: High-resolution video synthesis with latent diffusion models},
  author={Blattmann, Andreas and Rombach, Robin and Ling, Huan and Dockhorn, Tim and Kim, Seung Wook and Fidler, Sanja and Kreis, Karsten},
  booktitle={CVPR},
  year={2023}
}

@misc{wan2025wan,
  title         = {{Wan}: Open and Advanced Large-Scale Video Generative Models},
  author        = {{Team Wan} and Wang, Ang and Ai, Baole and Wen, Bin and Mao, Chaojie and Xie, Chen-Wei and Chen, Di and Yu, Feiwu and Zhao, Haiming and Yang, Jianxiao and Zeng, Jianyuan and Wang, Jiayu and Zhang, Jingfeng and Zhou, Jingren and Wang, Jinkai and Chen, Jixuan and Zhu, Kai and Zhao, Kang and Yan, Keyu and Huang, Lianghua and Feng, Mengyang and Zhang, Ningyi and Li, Pandeng and Wu, Pingyu and Chu, Ruihang and Feng, Ruili and Zhang, Shiwei and Sun, Siyang and Fang, Tao and Wang, Tianxing and Gui, Tianyi and Weng, Tingyu and Shen, Tong and Lin, Wei and others},
  year          = {2025},
  eprint        = {2503.20314},
  archivePrefix = {arXiv},
  primaryClass  = {cs.CV},
  doi           = {10.48550/arXiv.2503.20314},
  url           = {https://arxiv.org/abs/2503.20314}
}

@inproceedings{yang2025longlive,
  title     = {{LongLive}: Real-time Interactive Long Video Generation},
  author    = {Yang, Shuai and Huang, Wei and Chu, Ruihang and Xiao, Yicheng and Zhao, Yuyang and Wang, Xianbang and Li, Muyang and Xie, Enze and Chen, Ying-Cong and Lu, Yao and Han, Song and Chen, Yukang},
  booktitle = {The Fourteenth International Conference on Learning Representations},
  year      = {2026},
  url       = {https://openreview.net/forum?id=nCAODkpsPJ},
  eprint    = {2509.22622},
  archivePrefix = {arXiv},
  primaryClass  = {cs.CV}
}

@inproceedings{liu2026rollingforcing,
  title     = {Rolling Forcing: Autoregressive Long Video Diffusion in Real Time},
  author    = {Liu, Kunhao and Hu, Wenbo and Xu, Jiale and Shan, Ying and Lu, Shijian},
  booktitle = {The Fourteenth International Conference on Learning Representations},
  year      = {2026},
  url       = {https://openreview.net/forum?id=IAyzXjbfwo},
  eprint    = {2509.25161},
  archivePrefix = {arXiv},
  primaryClass  = {cs.CV}
}

@inproceedings{cui2025selfforcingpp,
  title     = {Self-Forcing++: Towards Minute-Scale High-Quality Video Generation},
  author    = {Cui, Justin and Wu, Jie and Li, Ming and Yang, Tao and Li, Xiaojie and Wang, Rui and Bai, Andrew and Ban, Yuanhao and Hsieh, Cho-Jui},
  booktitle = {The Fourteenth International Conference on Learning Representations},
  year      = {2026},
  url       = {https://openreview.net/forum?id=DzvPiqh23f}
}

@article{jin2024pyramidal,
  title={Pyramidal flow matching for efficient video generative modeling},
  author={Jin, Yang and Sun, Zhicheng and Li, Ningyuan and Xu, Kun and Jiang, Hao and Zhuang, Nan and Huang, Quzhe and Song, Yang and Mu, Yadong and Lin, Zhouchen},
  journal={arXiv preprint arXiv:2410.05954},
  year={2024}
}

@article{chen2024diffusion,
  title={Diffusion forcing: Next-token prediction meets full-sequence diffusion},
  author={Chen, Boyuan and Mart{\'\i} Mons{\'o}, Diego and Du, Yilun and Simchowitz, Max and Tedrake, Russ and Sitzmann, Vincent},
  journal={Advances in Neural Information Processing Systems},
  volume={37},
  pages={24081--24125},
  year={2024}
}

@inproceedings{lipman2022flow,
  title     = {Flow Matching for Generative Modeling},
  author    = {Lipman, Yaron and Chen, Ricky T. Q. and Ben-Hamu, Heli and Nickel, Maximilian and Le, Matthew},
  booktitle = {The Eleventh International Conference on Learning Representations},
  year      = {2023},
  url       = {https://openreview.net/forum?id=PqvMRDCJT9t}
}

@inproceedings{esser2024scaling,
  title     = {Scaling Rectified Flow Transformers for High-Resolution Image Synthesis},
  author    = {Esser, Patrick and Kulal, Sumith and Blattmann, Andreas and Entezari, Rahim and M{\"u}ller, Jonas and Saini, Harry and Levi, Yam and Lorenz, Dominik and Sauer, Axel and Boesel, Frederic and Podell, Dustin and Dockhorn, Tim and English, Zion and Rombach, Robin},
  booktitle = {Proceedings of the 41st International Conference on Machine Learning},
  pages     = {12606--12633},
  year      = {2024},
  volume    = {235},
  series    = {Proceedings of Machine Learning Research},
  publisher = {PMLR},
  url       = {https://proceedings.mlr.press/v235/esser24a.html}
}

@article{song2023consistency,
  title={Consistency Models},
  author={Song, Yang and Dhariwal, Prafulla and Chen, Mark and Sutskever, Ilya},
  journal={arXiv preprint arXiv:2303.01469},
  year={2023},
}

@article{song2023improved,
  title={Improved techniques for training consistency models},
  author={Song, Yang and Dhariwal, Prafulla},
  journal={arXiv preprint arXiv:2310.14189},
  year={2023}
}

@inproceedings{yin2024dmd,
  title     = {One-step Diffusion with Distribution Matching Distillation},
  author    = {Yin, Tianwei and Gharbi, Micha{\"e}l and Zhang, Richard and Shechtman, Eli and Durand, Fr{\'e}do and Freeman, William T. and Park, Taesung},
  booktitle = {Proceedings of the IEEE/CVF Conference on Computer Vision and Pattern Recognition (CVPR)},
  pages     = {6613--6623},
  year      = {2024},
  month     = {June},
  doi       = {10.1109/CVPR52733.2024.00632},
  url       = {https://openaccess.thecvf.com/content/CVPR2024/html/Yin_One-step_Diffusion_with_Distribution_Matching_Distillation_CVPR_2024_paper.html}
}

@inproceedings{yin2024dmd2,
  title     = {Improved Distribution Matching Distillation for Fast Image Synthesis},
  author    = {Yin, Tianwei and Gharbi, Micha{\"e}l and Park, Taesung and Zhang, Richard and Shechtman, Eli and Durand, Fr{\'e}do and Freeman, William T.},
  booktitle = {Advances in Neural Information Processing Systems},
  volume    = {37},
  pages     = {47455--47487},
  year      = {2024},
  editor    = {Globerson, A. and Mackey, L. and Belgrave, D. and Fan, A. and Paquet, U. and Tomczak, J. and Zhang, C.},
  publisher = {Curran Associates, Inc.},
  doi       = {10.52202/079017-1505},
  url       = {https://proceedings.neurips.cc/paper_files/paper/2024/file/54dcf25318f9de5a7a01f0a4125c541e-Paper-Conference.pdf}
}

@misc{nie2026tmd,
  title         = {Transition Matching Distillation for Fast Video Generation},
  author        = {Nie, Weili and Berner, Julius and Ma, Nanye and Liu, Chao and Xie, Saining and Vahdat, Arash},
  year          = {2026},
  eprint        = {2601.09881},
  archivePrefix = {arXiv},
  primaryClass  = {cs.CV},
  doi           = {10.48550/arXiv.2601.09881},
  url           = {https://arxiv.org/abs/2601.09881}
}

@misc{lu2025rewardforcing,
  title         = {Reward Forcing: Efficient Streaming Video Generation with Rewarded Distribution Matching Distillation},
  author        = {Lu, Yunhong and Zeng, Yanhong and Li, Haobo and Ouyang, Hao and Wang, Qiuyu and Cheng, Ka Leong and Zhu, Jiapeng and Cao, Hengyuan and Zhang, Zhipeng and Zhu, Xing and Shen, Yujun and Zhang, Min},
  year          = {2025},
  eprint        = {2512.04678},
  archivePrefix = {arXiv},
  primaryClass  = {cs.CV},
  doi           = {10.48550/arXiv.2512.04678},
  url           = {https://arxiv.org/abs/2512.04678}
}

@inproceedings{liu2026diagonaldistill,
  title     = {Streaming Autoregressive Video Generation via Diagonal Distillation},
  author    = {Liu, Jinxiu and Liu, Xuanming and Mei, Kangfu and Wen, Yandong and Yang, Ming-Hsuan and Liu, Weiyang},
  booktitle = {The Fourteenth International Conference on Learning Representations},
  year      = {2026},
  url       = {https://openreview.net/forum?id=X7YW6STzeL},
  eprint    = {2603.09488},
  archivePrefix = {arXiv},
  primaryClass  = {cs.CV}
}

@article{wang2023prolificdreamer,
  title={Prolificdreamer: High-fidelity and diverse text-to-3d generation with variational score distillation},
  author={Wang, Zhengyi and Lu, Cheng and Wang, Yikai and Bao, Fan and Li, Chongxuan and Su, Hang and Zhu, Jun},
  journal={Advances in neural information processing systems},
  volume={36},
  pages={8406--8441},
  year={2023}
}

@article{luo2023diff,
  title={Diff-instruct: A universal approach for transferring knowledge from pre-trained diffusion models},
  author={Luo, Weijian and Hu, Tianyang and Zhang, Shifeng and Sun, Jiacheng and Li, Zhenguo and Zhang, Zhihua},
  journal={Advances in Neural Information Processing Systems},
  volume={36},
  pages={76525--76546},
  year={2023}
}

@article{wang2024vidprom,
  title={Vidprom: A million-scale real prompt-gallery dataset for text-to-video diffusion models},
  author={Wang, Wenhao and Yang, Yi},
  journal={Advances in Neural Information Processing Systems},
  volume={37},
  pages={65618--65642},
  year={2024}
}

@inproceedings{huang2024vbench,
  title={Vbench: Comprehensive benchmark suite for video generative models},
  author={Huang, Ziqi and He, Yinan and Yu, Jiashuo and Zhang, Fan and Si, Chenyang and Jiang, Yuming and Zhang, Yuanhan and Wu, Tianxing and Jin, Qingyang and Chanpaisit, Nattapol and others},
  booktitle={Proceedings of the IEEE/CVF Conference on Computer Vision and Pattern Recognition},
  pages={21807--21818},
  year={2024}
}

@article{xu2024visionreward,
  title={Visionreward: Fine-grained multi-dimensional human preference learning for image and video generation},
  author={Xu, Jiazheng and Huang, Yu and Cheng, Jiale and Yang, Yuanming and Xu, Jiajun and Wang, Yuan and Duan, Wenbo and Yang, Shen and Jin, Qunlin and Li, Shurun and others},
  journal={arXiv preprint arXiv:2412.21059},
  year={2024}
}

@article{chen2025skyreels,
  title={Skyreels-v2: Infinite-length film generative model},
  author={Chen, Guibin and Lin, Dixuan and Yang, Jiangping and Lin, Chunze and Zhu, Junchen and Fan, Mingyuan and Zhang, Hao and Chen, Sheng and Chen, Zheng and Ma, Chengcheng and others},
  journal={arXiv preprint arXiv:2504.13074},
  year={2025}
}

@article{teng2025magi,
  title={MAGI-1: Autoregressive Video Generation at Scale},
  author={Teng, Hansi and Jia, Hongyu and Sun, Lei and Li, Lingzhi and Li, Maolin and Tang, Mingqiu and Han, Shuai and Zhang, Tianning and Zhang, WQ and Luo, Weifeng and others},
  journal={arXiv preprint arXiv:2505.13211},
  year={2025}
}

@misc{openai2024sora,
  author       = {Brooks, Tim and Peebles, Bill and Holmes, Connor and DePue, Will and Guo, Yufei and Jing, Li and Schnurr, David and Taylor, Joe and Luhman, Troy and Luhman, Eric and Ng, Clarence and Wang, Ricky and Ramesh, Aditya},
  title        = {Video generation models as world simulators},
  howpublished = {OpenAI Research Blog},
  year         = {2024},
  url          = {https://openai.com/index/video-generation-models-as-world-simulators/},
  note         = {Accessed: 2026-08-21}
}

@misc{seedance2026seedance,
  title         = {{Seedance 2.0}: Advancing Video Generation for World Complexity},
  author        = {{Team Seedance} and Chen, De and Chen, Liyang and Chen, Xin and Chen, Ying and Chen, Zhuo and Chen, Zhuowei and Cheng, Feng and Cheng, Tianheng and Cheng, Yufeng and Chi, Mojie and Chi, Xuyan and Cong, Jian and Cui, Qinpeng and Ding, Fei and Dong, Qide and others},
  year          = {2026},
  eprint        = {2604.14148},
  archivePrefix = {arXiv},
  primaryClass  = {cs.CV},
  doi           = {10.48550/arXiv.2604.14148},
  url           = {https://arxiv.org/abs/2604.14148}
}

@article{hu2021lora,
  title={Lora: Low-rank adaptation of large language models},
  author={Hu, Edward J and Shen, Yelong and Wallis, Phillip and Allen-Zhu, Zeyuan and Li, Yuanzhi and Wang, Shean and Wang, Lu and Chen, Weizhu},
  journal={arXiv preprint arXiv:2106.09685},
  year={2021}
}

@article{hacohen2024ltx,
  title={Ltx-video: Realtime video latent diffusion},
  author={HaCohen, Yoav and Chiprut, Nisan and Brazowski, Benny and Shalem, Daniel and Moshe, Dudu and Richardson, Eitan and Levin, Eran and Shiran, Guy and Zabari, Nir and Gordon, Ori and others},
  journal={arXiv preprint arXiv:2501.00103},
  year={2024}
}
\bibliographystyle{main}

\appendix
\clearpage

\section{More Implementation Details}
\label{training_details}

\paragraph{Noise schedule and model parameterization.}
Following Wan2.1, we use a flow-matching scheduler. For a sampled timestep $t\in[0,1000]$, the shifted noise level is
\[
    \sigma_t =
    \frac{k(t/1000)}{1+(k-1)(t/1000)},
\]
with shift factor $k=5$ for generator rollouts and for the DMD/GAN critic timestep sampling. The forward corruption process is
\[
    x_t=(1-\sigma_t)x_0+\sigma_t\epsilon,\qquad
    \epsilon\sim\mathcal{N}(0,I),
\]
and the flow-prediction target is $\epsilon-x_0$. During generation, the model predicts $v_\theta(x_t,t,c)$ and converts it to a clean latent estimate by $\hat{x}_0=x_t-\sigma_t v_\theta(x_t,t,c)$.

\paragraph{Training details of our method.} 

We first construct a dataset $\mathcal{D}_\text{Bi}$ consisting of about 3K samples generated by Wan ~\citep{wan2025wan} with the VidProM~\citep{wang2024vidprom} prompts. Next, we sample ODE trajectories from the autoregressive diffusion model to construct a ODE Slow-Fast dataset $\mathcal{D}_{\text{Slow-Fast}}$ with 3K samples. 
Throughout training, including the teacher forcing autoregressive diffusion model and both ODE-initialization variants, we use either $D_{\text{Bi}}$ or $\mathcal{D}_{\text{Slow-Fast}}$, both internally synthesized by the model, and using about the same prompts, thus guaranteeing that there is no gap in terms of data quality.  
Then, we perform causal ODE distillation on $\mathcal{D}_{\text{Slow-Fast}}$ for 1K steps via teacher forcing, conditioned on the corresponding clean data in $\mathcal{D}_\text{Bi}$. 
In this stage, the ODE student is initialized from the autoregressive diffusion teacher. Finally, we use this model as initialization for the standard asymmetric DMD, where $s_\text{real}$ is Wan2.1-14B, and $s_\text{fake}$ is Wan2.1-1.3B, strictly following the setting of Self Forcing~\citep{huang2026self} and Causal Forcing~\citep{zhu2026causal} to guarantee the fair comparison.

Throughout all training stages, we use a batch size of 64 and the Adam optimizer with a learning rate of $2\times10^{-6}$, $\beta_1=0$, and $\beta_2=0.999$, while keeping all other settings identical to Causal Forcing. During inference, we adopt a 4-step sampling scheme using shared timesteps from the causal ODE initialization and asymmetric DMD: Slow-LoRA is assigned to timesteps $1$ and $0.937$, while Fast-LoRA takes charge of timesteps $0.833$ and $0.625$.
Both adapters use a LoRA rank of 32 and are inserted into all linear layers of and $\alpha$ of 128 of the diffusion transformer. 
Training is performed with a global batch size of $1$, mixed-precision (\texttt{bf16}), and gradient clipping of $1.0$. We train for a total of $60$ steps, which corresponds to effectively training the single data sample for $60$ epochs, and enable 8-bit Adam and quantized weight loading to reduce memory footprint. 

\section{Analysis Details}
\paragraph{Trajectory Curvature.}
This section provides additional details on the trajectory sampling and robustness analysis used to compute the trajectory curvature reported in Section
~\ref{suboptimal}. We evaluate 100 denoising trajectories generated by Wan2.1-T2V-1.3B~\citep{wan2025wan}, covering diverse prompts with different subjects, scene structures, and motion patterns. For each prompt, we use the standard 50-step denoising process with a noise schedule shift of 8 and classifier-free guidance scale of 6.
To eliminate randomness introduced by different initial noise realizations, we use one deterministic seed for each prompt.

Cross the 100 trajectories, the curvature is strongly concentrated in the high-noise regime. In particular, the per-prompt estimates indicate that $91.82\%\pm0.15\%$ of the total curvature mass is accumulated at $t\geq0.9$, where the uncertainty is reported as the standard error of the mean (SEM). The corresponding 95\% bootstrap confidence interval is $[91.53,92.11]\%$. Moreover, the ratio between the curvature accumulated in the high-noise and mid-noise regimes is $32.4\pm0.6$, with a 95\% bootstrap confidence interval of $[31.2,33.6]$. These results indicate that the strong trajectory bending observed in the early denoising stage is not caused by a small number of atypical prompts, but is consistently observed across the sampled trajectories.

A temporal-difference version of the same metric, which removes static appearance and emphasizes motion structure, still places $87.9\%$ of the curvature mass at $t\ge0.9$ with a high/mid ratio of $18.6$.

\paragraph{One-Sample Selection Criteria.}
We construct six representative samples along three dimensions: \textit{motion dynamics}, \textit{visual complexity}, and \textit{domain specificity}. Each dimension contains two contrasting samples under consistent video settings, camera trajectories, subject scales, and prompt structures. Specifically, High Motion ans Low Motion distinguish substantial from minimal temporal dynamics; Complex Textures and Simple Geometry contrast dense dynamic textures with clean geometric scenes; and Open Domain and Specific Domain contrast broadly recognizable content with relatively rare or specialized content. This controlled categorization allows us to isolate different sources of generation difficulty while covering diverse visual scenarios.

The generation configuration and prompt structure are kept consistent across all samples. In particular, the videos are generated using Wan2.1-T2V-1.3B with the same sampling configuration described above. We also normalize the prompt structure, subject scale, and camera trajectory description across samples so that the selected dimensions primarily capture differences in motion, visual complexity, and domain specificity rather than differences in generation settings. The detailed prompts are provided in Appendix~\ref{prompt}.

\section{User Study}
\label{app:user-study}

\begin{figure}[ht]
\centering
\includegraphics[width=\columnwidth]{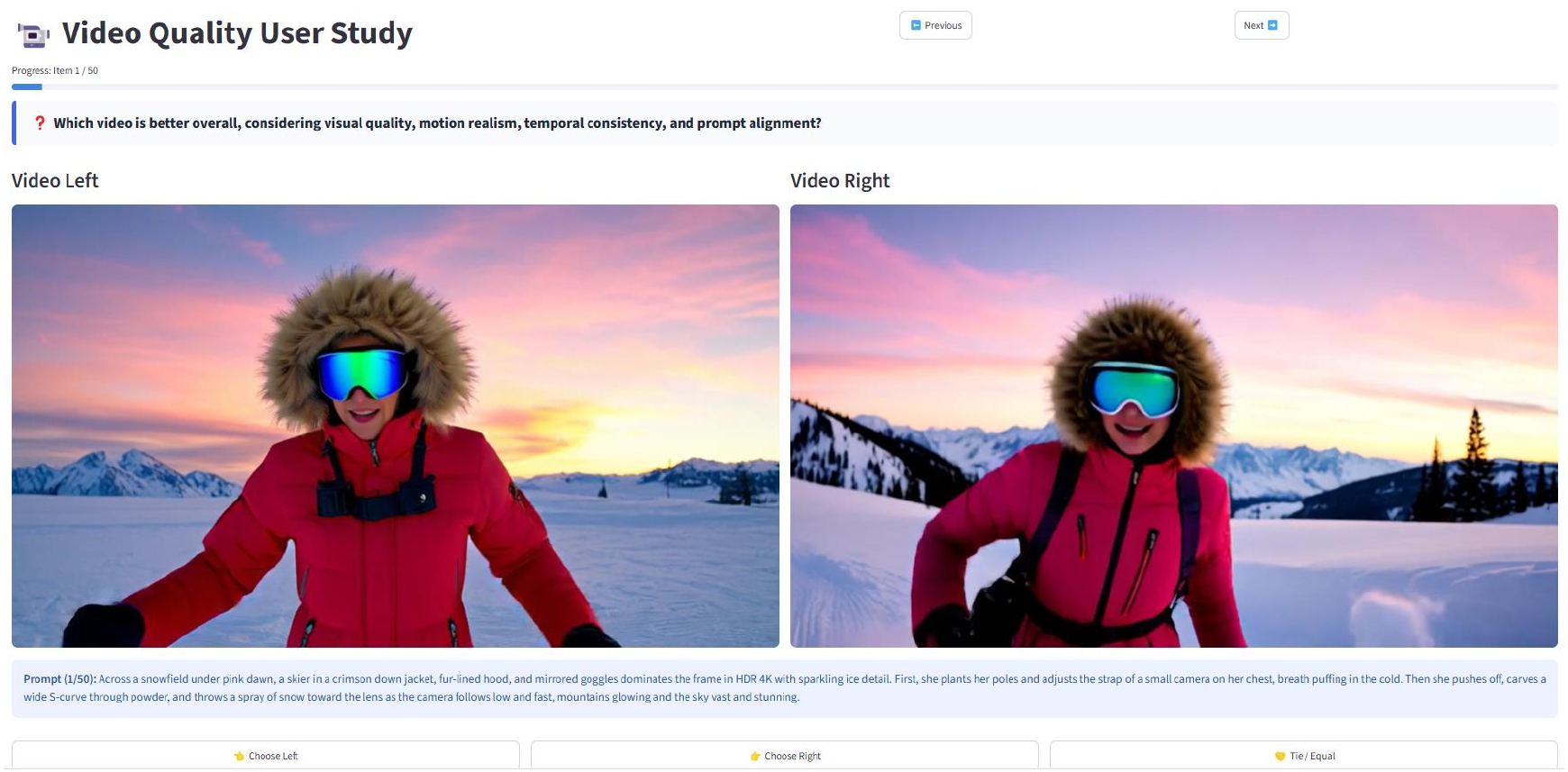}
\caption{User Study platform.}
\label{user_appendix}
\end{figure}

To collect human feedback, we conducted a web-based pairwise user study,
with the evaluation interface shown in Figure~\ref{user_appendix}. The
platform automatically randomizes the order and spatial placement of model
outputs to reduce potential presentation bias.

We compare \method{} against two baselines, Self Forcing and Causal
Forcing. For each comparison, we randomly sampled 50 text prompts and
generated the corresponding video pairs. All prompts were fixed before the
evaluation, and each video pair was independently evaluated by three human
annotators.

The two videos were displayed side-by-side and played continuously in a
loop. Annotators were asked to select the preferred video or indicate a tie,
considering overall visual quality, temporal consistency, motion realism,
and prompt adherence. The evaluation question was:

\begin{quote}
\emph{``Which video is better overall, considering visual quality, motion
realism, temporal consistency, and prompt alignment?''}
\end{quote}

Each comparison therefore contains $50 \times 3 = 150$ individual ratings,
resulting in 300 ratings across the two comparisons. The aggregated voting
results are reported in Table~\ref{user} in the main paper.

\section{More Abalation Study}
\paragraph{Sensitivity to the Boundary Timestep $\kappa$.}
\begin{wrapfigure}{r}{0.5\textwidth}
\centering
\includegraphics[width=\linewidth]{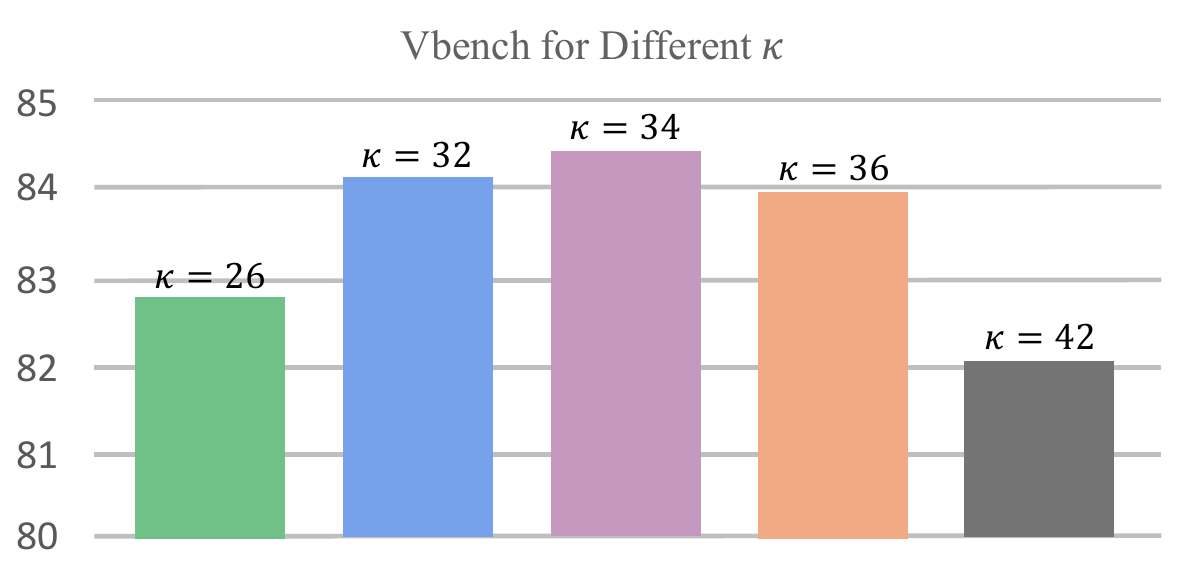}
\caption{Impact of different $\kappa$.}
\label{abla2}
\end{wrapfigure}
Based on the trajectory-curvature analysis in Figure~\ref{fig11}(a), the relative velocity deviation rapidly decreases and becomes nearly stable around step 34 ($t\approx0.9$), naturally dividing the diffusion trajectory into semantic synthesis and detail refinement stages.
To validate this observation, we evaluate different transition points $\kappa\in\{26,32,34,36,42\}$. As shown in Figure~\ref{abla2}, the best performance is consistently achieved at $\kappa=34$, while moving the boundary toward either earlier or later timesteps leads to noticeable performance degradation. These results support our trajectory analysis and demonstrate that properly balancing the two optimization phases is crucial for achieving optimal generation quality.

\section{Additional Qualitative Results}
\begin{figure*}[t]
\centering
\includegraphics[width=1\textwidth]{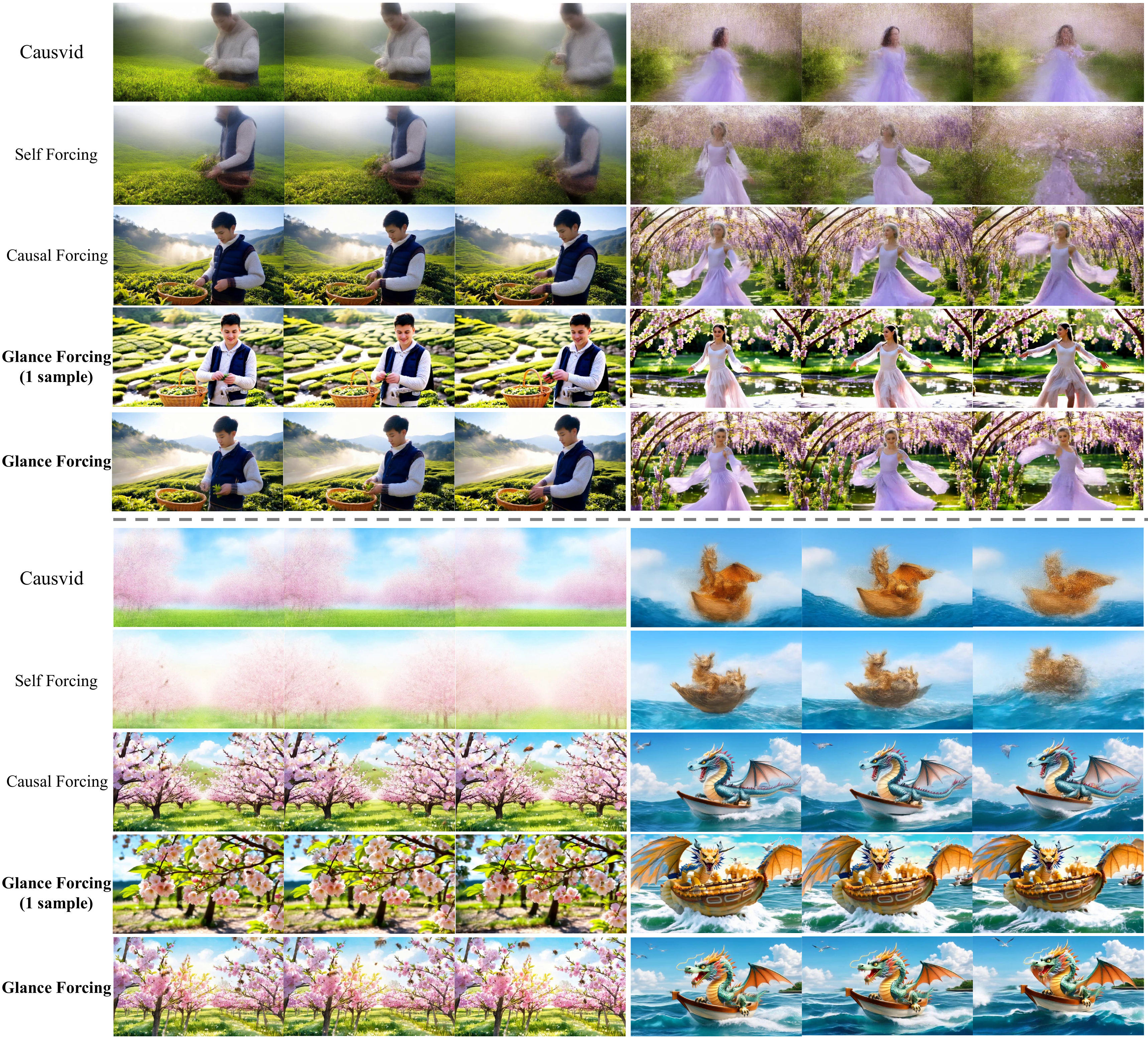}
\caption{Qualitative comparisons with existing methods for Initialized stage.}
\label{appendix_ode}
\end{figure*}

\begin{figure*}[t]
\centering
\includegraphics[width=1\textwidth]{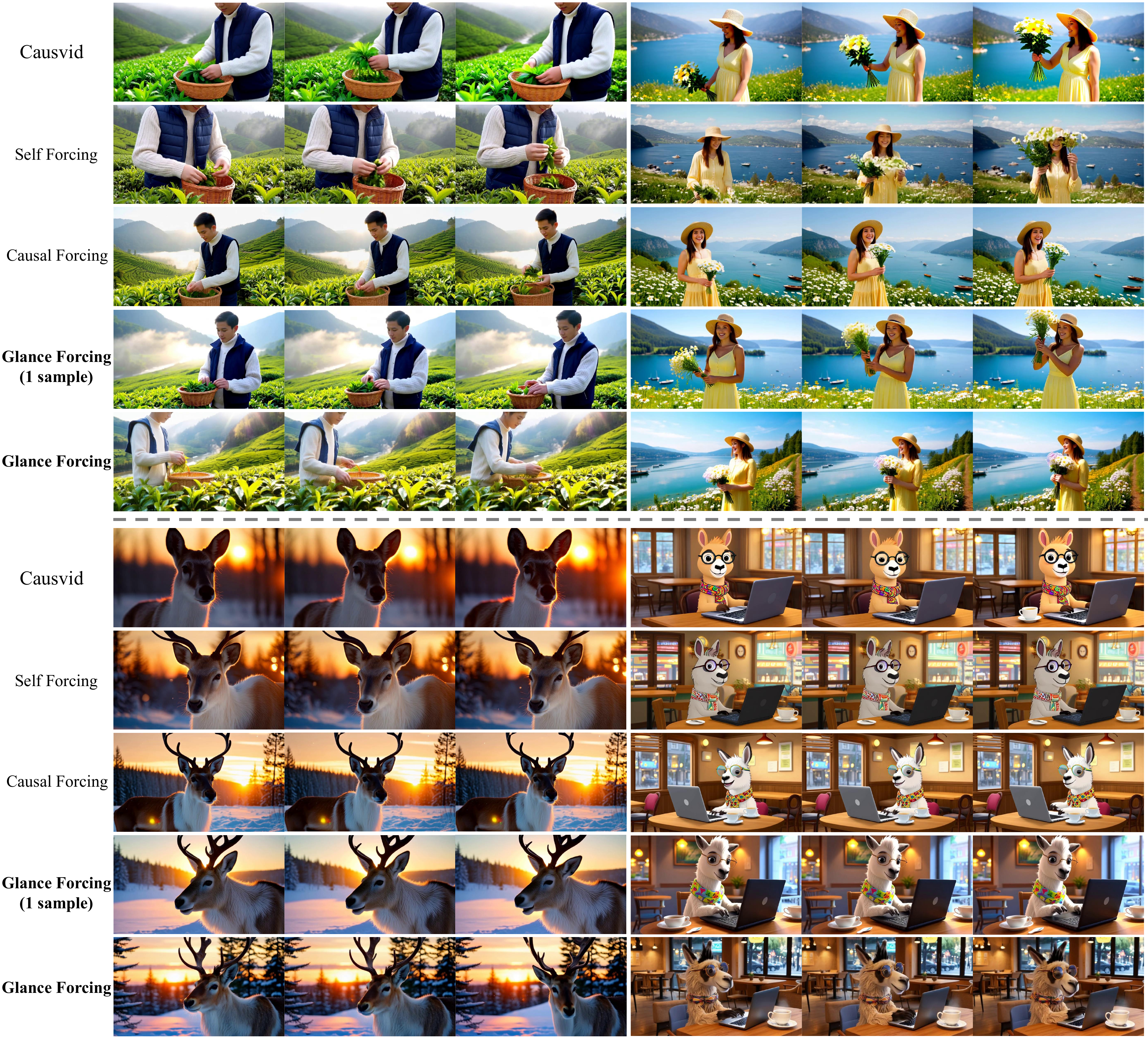}
\caption{Qualitative comparisons with existing methods for distilled stage.}
\label{appendix_dmd}
\end{figure*}

We provide additional qualitative results on five new prompts, covering all six models in Table~\ref{table1}. For each prompt, every model generates five videos with the same random seed.
Figure~\ref{appendix_ode} further presents a qualitative comparison between our method and other approaches during the ODE initialization phase, while Figure~\ref{appendix_dmd} illustrates additional qualitative comparisons during the DMD distillation phase.


\section{Prompts}
\label{prompt}
\textbf{The prompt of the video of Table 4:}

1."A cinematic 5-second medium-wide shot of a sleek red sports car racing through a winding mountain road at sunset, occupying approximately 50–60\% of the frame. The car rapidly accelerates toward a sharp hairpin turn and performs a fast controlled drift, with the wheels spinning quickly, thick tire smoke expanding behind the vehicle, and small particles of dust and gravel scattering from the road. The vehicle moves continuously across the scene with strong physical momentum and noticeable motion blur. The camera follows the same smooth trajectory throughout the shot: a slow forward tracking movement combined with a very subtle left-to-right pan, maintaining the car near the center of the frame. The surrounding mountains and roadside barriers exhibit clear parallax caused by the rapid vehicle motion. Warm sunlight creates realistic highlights and shadows on the glossy red bodywork. No cuts, no scene transitions, continuous coherent motion, realistic vehicle dynamics, photorealistic cinematic footage, highly detailed, 4K."

2."A cinematic 5-second medium-wide shot of a young woman standing outdoors in a quiet natural environment, occupying approximately 50–60\% of the frame. The woman remains almost completely still throughout the shot, maintaining the same relaxed posture and calm facial expression. Only extremely subtle physical movements occur: gentle breathing, one natural blink, and a few loose strands of hair moving slightly in a soft breeze. Her clothing remains almost stationary with only minimal fabric movement. The camera follows the same smooth trajectory throughout the shot: a slow forward tracking movement combined with a very subtle left-to-right pan, maintaining the woman near the center of the frame. The background remains stable with only barely perceptible movement in distant vegetation. Soft natural daylight produces consistent shadows and realistic skin tones. No sudden movements, no walking, no head turning, no cuts, no scene transitions, continuous temporal consistency, photorealistic cinematic footage, highly detailed, 4K."

3."A cinematic 5-second medium-wide shot of a crowded night market street in Tokyo, with the central group of pedestrians and surrounding market stalls occupying approximately 50–60\% of the frame. Numerous people continuously walk through the scene while colorful signs, umbrellas, food stalls, wet pavement, steam, smoke, and water puddles create dense spatial textures. Light rain produces small ripples across the puddles, while reflections of red, blue, and orange lights continuously distort across the wet ground. Steam rises from food stalls and slowly mixes with drifting smoke, while clothing and umbrellas move naturally as people walk. The camera follows the same smooth trajectory throughout the shot: a slow forward tracking movement combined with a very subtle left-to-right pan, maintaining the main crowd near the center of the frame. Strong local motion occurs simultaneously across many different surfaces and objects. Neon lights create complex reflections and highlights across wet materials. No cuts, no scene transitions, continuous fluid and crowd motion, rich spatial detail, photorealistic cinematic footage, highly detailed, 4K."

4."A cinematic 5-second medium-wide shot of a single glossy chrome cube floating in a clean minimalist studio environment, with the cube occupying approximately 50–60\% of the frame. The cube has perfectly straight edges, six flat square faces, and a smooth reflective metallic surface. It continuously rotates around its vertical axis at a constant and moderate speed while remaining centered in the scene. The surrounding environment contains no other objects, people, textures, patterns, or visual clutter. The camera follows the same smooth trajectory throughout the shot: a slow forward tracking movement combined with a very subtle left-to-right pan, maintaining the cube near the center of the frame. Soft studio lights create stable reflections that move smoothly across the cube's metallic surface as it rotates. The background remains completely clean and uniform. No deformation, no additional objects, no cuts, no scene transitions, continuous rigid motion, simple geometric structure, photorealistic cinematic footage, highly detailed, 4K."

5."A cinematic 5-second medium-wide shot of a vast green valley under a bright blue sky, with the main landscape occupying approximately 50–60\% of the frame. Rolling hills extend toward the distant horizon, with a narrow river flowing through the valley and tall grass covering the foreground. Soft white clouds slowly drift across the sky while the grass gently sways in a light breeze and subtle highlights shimmer across the surface of the river. The natural environment contains familiar objects and common visual concepts without unusual or specialized elements. The camera follows the same smooth trajectory throughout the shot: a slow forward tracking movement combined with a very subtle left-to-right pan, gradually revealing the valley while maintaining the main landscape near the center of the frame. Warm sunlight produces consistent natural shadows and atmospheric depth across the hills. No sudden movements, no cuts, no scene transitions, continuous environmental motion, natural spatial composition, photorealistic cinematic footage, highly detailed, 4K."

6."A cinematic 5-second medium-wide shot of a single bioluminescent deep-sea jellyfish floating in a completely dark underwater environment, with the jellyfish occupying approximately 50–60\% of the frame. The translucent jellyfish has a glowing blue and violet bell and several long delicate tentacles extending beneath its body. It slowly contracts and expands its bell while drifting gently forward, causing the tentacles to deform and sway naturally with the surrounding water. Tiny suspended particles move slowly through the dark water at different depths, creating subtle spatial depth. The camera follows the same smooth trajectory throughout the shot: a slow forward tracking movement combined with a very subtle left-to-right pan, maintaining the jellyfish near the center of the frame. The bioluminescent body produces soft volumetric illumination and realistic scattering through the surrounding water. Translucency, refraction, subsurface scattering, and fluid motion remain visually consistent throughout the shot. No cuts, no scene transitions, continuous physical motion, photorealistic cinematic footage, highly detailed, 4K."

\textbf{The prompt of the video of Figure 6:}

1."Along a rainforest suspension bridge, a biologist in a yellow poncho and field vest fills the foreground in HDR 4K with wet leaves and bright insects. First, she checks a notebook, raises binoculars, and freezes as a toucan calls from the canopy. Then she lowers the binoculars, points excitedly, and walks forward carefully while the camera follows, revealing dripping vines, sunlit mist, and a deep green world that feels endless and alive. far from any noise. with the subject filling frame."

2."At a rooftop garden above Seoul at night, a violinist in a black velvet coat and sparkling hairpin stands prominent in ultra-realistic 4K with city lights bokeh. First, she tightens one string and holds still as a breeze lifts her coat hem, distant traffic humming. Then she draws the bow into a bright phrase and takes two steps forward, the camera drifting closer as skyscrapers glitter behind her and lanterns glow, textures rich and the mood intensely alive. while the skyline shimmers softly."

\textbf{The prompt of the video of Figure 7:}

1."A scenic photograph capturing the moment a steam train departs from the Glenfinnan Viaduct, a historic railway bridge in Scotland. The train moves gracefully over the arch-covered viaduct, its smoke billowing into the air. The landscape is lush with greenery, and towering rocky mountains frame the scene, creating a picturesque backdrop. The sky is a clear, bright blue with the sun shining down, casting a warm glow on the train and the surrounding scenery. The viaduct itself is a striking feature, with intricate ironwork and a verdant setting. The photo has a classic, nostalgic feel, emphasizing the natural beauty and historical charm of the location. A wide-angle shot from a slightly elevated angle, capturing both the train and the expansive landscape."

2."A cinematic closeup and detailed portrait of a reindeer standing in a snowy forest at sunset. The lighting is gorgeous and soft, with a golden backlight creating a warm and dreamy effect. Soft bokeh and lens flares add a magical touch, enhancing the cinematic quality of the image. The reindeer has a gentle expression, its fur glistening in the fading light. The background features a serene snowy landscape with tall trees silhouetted against the orange and pink hues of the setting sun. The color grade is rich and magical, capturing the essence of a winter wonderland at twilight. A close-up shot from a slightly elevated angle."

\textbf{The prompt of the video of Figure 11:}

1."Early spring in a mountain tea valley, a young man in a cream knit sweater and navy vest stands close to camera, ultra‑clean 4K. He plucks tender leaves into a woven basket, then wipes dew from his fingers as sun breaks through thin fog. The camera floats low over bright rows of tea, then rises to reveal layered ridges and a river ribbon shining below. Clothing stitches, leaf veins, and mist particles remain razor sharp and natural."

2."A realistic style paper origami dragon riding a boat through waves, with intricate folds and textures. The dragon has a fierce expression, its eyes glowing with intensity, and its scales shimmering in the sunlight. It is perched on the edge of the boat, wings partially spread, ready to take flight. The boat bobs up and down with the waves, creating a dynamic motion. The water is choppy, with ripples and splashes around the boat, adding to the sense of movement. The background features a clear blue sky with fluffy clouds, and a few seagulls flying overhead. A mid-shot capturing the dragon's powerful stance and the boat's motion."

3."Early spring in a mountain tea valley, a young man in a cream knit sweater and navy vest stands close to camera, ultra‑clean 4K. He plucks tender leaves into a woven basket, then wipes dew from his fingers as sun breaks through thin fog. The camera floats low over bright rows of tea, then rises to reveal layered ridges and a river ribbon shining below. Clothing stitches, leaf veins, and mist particles remain razor sharp and natural."

4."A realistic style paper origami dragon riding a boat through waves, with intricate folds and textures. The dragon has a fierce expression, its eyes glowing with intensity, and its scales shimmering in the sunlight. It is perched on the edge of the boat, wings partially spread, ready to take flight. The boat bobs up and down with the waves, creating a dynamic motion. The water is choppy, with ripples and splashes around the boat, adding to the sense of movement. The background features a clear blue sky with fluffy clouds, and a few seagulls flying overhead. A mid-shot capturing the dragon's powerful stance and the boat's motion."

\textbf{The prompt of the video of Figure 12:}

1."Early spring in a mountain tea valley, a young man in a cream knit sweater and navy vest stands close to camera, ultra‑clean 4K. He plucks tender leaves into a woven basket, then wipes dew from his fingers as sun breaks through thin fog. The camera floats low over bright rows of tea, then rises to reveal layered ridges and a river ribbon shining below. Clothing stitches, leaf veins, and mist particles remain razor sharp and natural."

2."A cinematic closeup and detailed portrait of a reindeer standing in a snowy forest at sunset. The lighting is gorgeous and soft, with a golden backlight creating a warm and dreamy effect. Soft bokeh and lens flares add a magical touch, enhancing the cinematic quality of the image. The reindeer has a gentle expression, its fur glistening in the fading light. The background features a serene snowy landscape with tall trees silhouetted against the orange and pink hues of the setting sun. The color grade is rich and magical, capturing the essence of a winter wonderland at twilight. A close-up shot from a slightly elevated angle."

3."A digital illustration in a whimsical cartoon style of a llama coding and typing on his laptop in a cozy cafe. The llama has a friendly expression, with large, expressive eyes and a gentle smile. It wears a colorful patterned scarf and a pair of round glasses perched on its nose. The cafe setting includes a wooden table, a few chairs, and a window with a view of a bustling street outside. The background is filled with the soft glow of ambient lighting and hints of other patrons. The llama's fingers dance over the keyboard, with a cup of steaming coffee nearby. A close-up shot from a slightly elevated angle, capturing the llama's focused and engaged posture."

4."In spring on a hillside above a bright lake, a woman in a light yellow dress and straw hat dominates the frame in cinematic 4K. She picks wildflowers, then lifts the bouquet toward the lens as the camera glides closer to her joyful expression. Behind her, water shimmers, small boats drift, and mountains soften into blue haze. Sharp textures, true tones, and gentle film grain create a warm, uplifting beauty."

\end{document}